\documentclass[sn-mathphys,Numbered]{sn-jnl}

\usepackage{todonotes}%
\usepackage{graphicx}%
\usepackage{multirow}%
\usepackage{amsmath,amssymb,amsfonts}%
\usepackage{amsthm}%
\usepackage{mathrsfs}%
\usepackage[title]{appendix}%
\usepackage{xcolor}%
\usepackage{textcomp}%
\usepackage{manyfoot}%
\usepackage{booktabs}%
\usepackage{algorithm}%
\usepackage{enumitem}%
\usepackage{algorithmicx}%
\usepackage{algpseudocode}%
\usepackage{listings}%
\usepackage{xspace}

\def\paramx {\taskize{A}}
\def\paramy {\taskize{B}}

\newcommand{\taskize}[1] {\ensuremath{\scalebox{0.85}{\textsf{#1}}}}

\def\ExiTxt {{existence}}

\def\AbseTxt {absence}

\def\InitTxt {Init}

\def\ResExTxt {responded\_existence}
\def\RespTxt {response}

\def\AltRespTxtShort {Alt.Response}
\def\ChaRespTxt {ChainResponse}
\def\PrecTxt {precedence}

\newcommand{\Exi}[2] {\ensuremath{\textsc{\ExiTxt}(#1,#2)}}

\newcommand{\Abse}[2] {\ensuremath{\textsc{\AbseTxt}(#1,#2)}}

\newcommand{\Ini}[1] {\ensuremath{\textsc{\InitTxt}(#1)}}

\newcommand{\ResEx}[2] {\ensuremath{\textsc{\ResExTxt(#1,#2)}}}
\newcommand{\Resp}[2] {\ensuremath{\textsc{\RespTxt(#1,#2)}}}

\newcommand{\AltRespShort}[2] {\ensuremath{\textsc{\AltRespTxtShort}(#1,#2)}}

\newcommand{\ChaRes}[2] {\ensuremath{\textsc{\ChaRespTxt}(#1,#2)}}
\newcommand{\Prec}[2] {\ensuremath{{\textsc{\PrecTxt}(#1,#2)}}}

\newcommand{\activity}[1]{\taskize{#1}}
\newcommand{\act}[1]{\textnormal{\textsf{#1}}}
\newcommand{\attr}[1]{\textit{#1}}

\newcommand{\hr}{\emph{hit rate}\xspace}

\newcommand{\declare}{\textsc{Declare}\xspace}
\newcommand{\bk}{\emph{temporal background knowledge}\xspace}

\newcommand{\distmetrics}{\emph{distance}\xspace}
\newcommand{\implmetrics}{\emph{implausibility}\xspace}
\newcommand{\divemetrics}{\emph{diversity}\xspace}
\newcommand{\sparsity}{\emph{sparsity}\xspace}

\newcommand{\runtime}{\emph{runtime}\xspace}
\newcommand{\tracefit}{\emph{trace fitness}\xspace}

\newcommand{\boso}{\textnormal{\textsf{BOSO}}\xspace}
\newcommand{\aoso}{\textnormal{\textsf{AOSO}}\xspace}
\newcommand{\bomo}{\textnormal{\textsf{BOMO}}\xspace}
\newcommand{\aomo}{\textnormal{\textsf{AOMO}}\xspace}

\begin{document}

\title[Article Title]{Guiding the generation of counterfactual explanations through temporal background knowledge for Predictive Process Monitoring}

\author*[1,2]{\fnm{Andrei} \sur{Buliga}}\email{abuliga@fbk.eu}

\author[3]{\fnm{Chiara} \sur{Di Francescomarino}}

\author[1]{\fnm{Chiara} \sur{Ghidini}}
\author[1]{\fnm{Ivan} \sur{Donadello}}
\author[1]{\fnm{Fabrizio} \sur{Maria Maggi}}

\affil[1]{\orgname{Free University of Bozen-Bolzano}, \orgaddress{\city{Bolzano},\country{Italy}}}

\affil[2]{\orgname{Fondazione Bruno Kessler}, \orgaddress{\city{Trento} \country{Italy}}}

\affil[3]{\orgname{University of Trento}, \orgaddress{\city{Trento},\country{Italy}}}


\abstract{Counterfactual explanations suggest what should be different in the input instance
to change the outcome of an AI system. 
When dealing with counterfactual explanations in the field of Predictive Process Monitoring, however, control flow relationships among events have to be carefully considered. A counterfactual, indeed, should not violate control flow relationships among activities (\emph{temporal background knowledge}). 
Within the field of Explainability in Predictive Process Monitoring, there have been a series of works regarding counterfactual explanations for outcome-based predictions. However, none of them consider the inclusion of temporal background knowledge when generating these counterfactuals. 

In this work, we adapt state-of-the-art techniques for counterfactual generation in the domain of XAI that are based on genetic algorithms to consider a series of temporal constraints at runtime. 
We assume that this temporal background knowledge is given, and we adapt the fitness function, as well as the crossover and mutation operators, to maintain the satisfaction of the constraints. 
The proposed methods are evaluated with respect to state-of-the-art genetic algorithms for counterfactual generation and the results are presented. We showcase that the inclusion of temporal background knowledge allows the generation of counterfactuals more conformant to the temporal background knowledge, without however losing in terms of the counterfactual traditional quality metrics.
}


\maketitle

\section{Introduction}\label{sec:introduction}
A branch of Process Mining known as Predictive Process Monitoring (PPM) \cite{chiarappm1} deals with making predictions regarding the continuation of partially executed process instances using a historical record of past complete process executions.
State-of-the-art efforts in PPM have focused on delivering accurate predictive models through the application of ensemble learning and deep learning techniques~\cite{tax2017predictive,teinemaa2019outcome}. Due to the inherent complexity of these models, commonly referred to as \emph{black-box} models, they are often challenging for users to comprehend. For this reason, eXplainable Artificial Intelligence (XAI) techniques have been recently adopted in PPM to help interpret their predictions and foster adoption of these advanced predictive models~\cite{stierle2021bringing,rizzi2020lime,created,dice4el}.%

One example of such XAI techniques are counterfactual explanations, which suggest what should be different in the input instance to change the outcome of a prediction returned by a black-box predictive model. Most state-of-the-art counterfactual generation techniques in XAI are based around optimisation techniques~\cite{guidotti2022counterfactual}. Such techniques find the minimum change to the input that flips the prediction. 
Although the XAI literature has explored the utilization of GAs for the generation of counterfactuals, no clear mapping exists to go from counterfactual desiderata to concrete optimisation objectives. We thus contribute to the literature by providing a systematic mapping from the counterfactual explanation desiderata defined in~\cite{verma2020counterfactual} to concrete optimization objectives for GA-based approaches for counterfactual generation.

Counterfactual explanations are essential for providing alternatives to achieve a certain outcome (\emph{outcome-based predictions}) in the PPM domain. They can guide users in modifying the process instance to reach a desired outcome by assisting them in understanding how various activities or attributes may affect the process instance's outcome.
When dealing with counterfactual explanations in the PPM field, however, process constraints, as control flow relationships among events have to be carefully considered. A counterfactual, indeed, should not violate such process constraints.  
For instance, suppose a counterfactual suggests to a customer that, to be accepted, a loan application submission has to be performed before the loan application is actually filled in. This scenario would not benefit the customer as it goes against the logical sequence of events, violating certain temporal constraints and generating infeasible counterfactuals.

State-of-the-art methods for counterfactuals 
in PPM either utilize a gradient-based search, incorporating terms to maintain generated counterfactual feasibility~\cite{dice4el}, or employ genetic algorithms (GAs)~\cite{loreley,created}, preserving feasibility by treating the entire process execution as a single feature~\cite{loreley} or by building counterfactuals in a state-wise manner to determine their likelihood based on the observed data~\cite{created}. However, none of the approaches so far explicitly incorporate these temporal process constraints, that we can assume to know about the process (\bk), at runtime to ensure the feasibility of the generated counterfactual explanations. 
In this work, inspired by the process-aware dimension of the evaluation framework introduced in~\cite{caisepaper}, we adapt XAI GA-based state-of-the-art techniques for the counterfactual generation to the PPM scenario by explicitly incorporating \bk. To this aim, we adapt the fitness function, as well as the crossover and mutation operators, to incorporate the \bk at runtime. We investigate both the single and multi-objective formulation of the GA problem.
We compare the proposed adapted methods 
to state-of-the-art methods using an extensive evaluation protocol over different
real-life datasets. The results suggest that the approaches that include the \bk can produce better counterfactuals explanations, especially in terms of conformance with the provided \bk, while still maintaining or improving traditional counterfactual quality metrics.

The main contributions of the work are:
\begin{itemize}[label=$\bullet$, topsep=1pt]
    \item The first systematic mapping from counterfactual explanation desiderata to optimisation objectives for GAs. Indeed, although the XAI literature has explored the usage of GAs for the generation of counterfactuals, the concrete optimisation objectives used in these approaches have never been systematically mapped to the desiderata that a counterfactual should have~\cite{verma2020counterfactual,guidotti2022counterfactual}.
    \item The  formulation of a multi-objective optimisation problem for the generation of counterfactuals in the field of PPM. Indeed, although both single and multi-objective approaches have been explored in the XAI literature, concluding that multi-objective allows for obtaining better results~\cite{moc}, counterfactual generation methods based on GAs in PPM focus only on the single-objective optimisation case.
    \item A GA-based optimisation approach for the generation of counterfactual explanations for PPM that incorporate \bk at runtime.
    \item An extensive evaluation comparing the proposed approaches with state-of-the-art techniques over different real-life datasets.
\end{itemize}

The following section (Section~\ref{sec:motivating}) delves into a motivating example to highlight the importance of such contributions.  Section~\ref{sec:background} introduces the needed background to grasp the contents of the paper, Section~\ref{sec:relwork} discusses related work, Section~\ref{sec:approach} highlights the contributions of this work, while Section~\ref{sec:eval} introduces the evaluation and experimental settings. The analysis of the results is reported 
in Section~\ref{sec:results}, while the conclusions are outlined
in Section~\ref{sec:conclusions}.

\section{Motivating example}
\label{sec:motivating}
\begin{figure}[ht]
\centering
  \includegraphics[width=\textwidth,height=7cm]{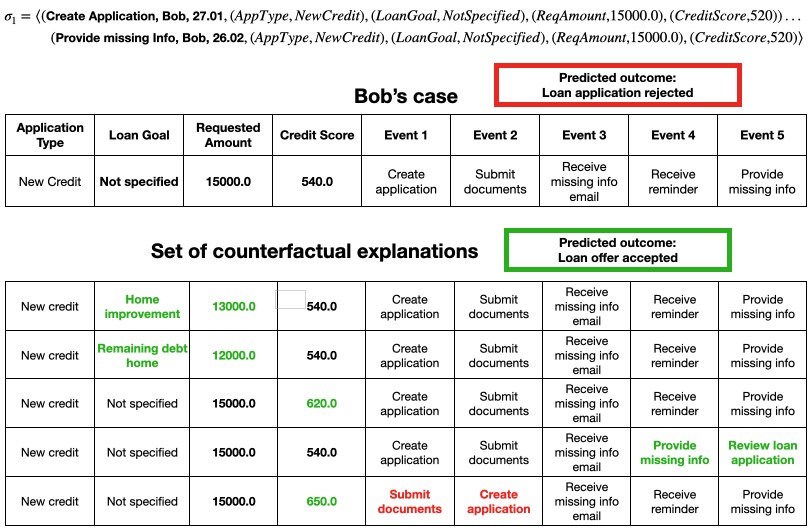}
  \caption{Example of an extract from an event long $\sigma_1$ and a set of counterfactual explanations in Predictive Process monitoring.}
  \vspace{-0.4cm}
  \label{fig:motivating_example}
\end{figure}

To illustrate a motivating scenario, we consider Bob, a customer of a bank that has applied for a loan and is waiting for a decision from the bank. Bob's application is shown in the upper part of \figurename~\ref{fig:motivating_example}. 
He has a credit score of 540 euros and has applied for a new credit application without specifying the goal of the loan. He requested an amount of $15\,000$ euros. After he has created the application (event1  \act{Create application}), he submits the documents for the application (\act{Submit documents}). He then receives an email from the bank asking for some missing information in the application request (\act{Receive missing info email}) and, after $30$ days, a reminder email, since he forgot to answer to the first one (\act{Receive reminder}). Only at that point in time, Bob provides an answer with the missing information to the bank (\act{Provide missing info}).

At some point during the processing of Bob's case, the bank returns, saying that his loan application has been rejected. Bob would like to know what changes he needs to make to get his loan application accepted. If we consider Bob's case as the original query instance, 
a possible set of counterfactual explanations that Bob could obtain is the one reported in the bottom part of \figurename~\ref{fig:motivating_example}.

This set of counterfactuals offers Bob a series of different feasible alternatives: (i) Bob could specify the loan goal and lower the requested amount, (ii) Bob could improve his credit score while keeping the same requested amount, or (iii) Bob could immediately answer to the bank's email by providing the missing info (\act{Provide missing info}) and then review his Loan application (\act{Review Loan application}). 
Among the returned counterfactuals, however, we also find a fourth alternative, highlighted in red in \figurename~\ref{fig:motivating_example}, 
suggesting that Bob should submit the documents for the application 
even before the application was created by Bob, something that is impossible to do. This counterfactual example shows a potential problem of traditional counterfactual generation approaches. 
When counterfactual generation techniques are simply applied to Predictive Process Monitoring in an out-of-the-box manner, they are not aware of the temporal relations between certain features, such as the control-flow activities that have certain constraints on their sequentiality. This highlights the importance of adapting counterfactual generation techniques to the field of Predictive Process Monitoring.

\section{Background and Preliminaries}\label{sec:background}
In this section, we introduce the main concepts useful to understand the remainder of the paper: event logs and the \textsc{Declare} language, as well as counterfactual explanations and genetic algorithms.

\subsection{Event Logs}
An \emph{event log} $\mathcal{L}$ is a set of traces (a.k.a.~cases), each representing one execution of a process.
A trace $\sigma= \langle e_{1}, e_{2}, \ldots e_{n} \rangle$ consists of a sequence of \emph{events} $e_i$, each referring to the execution of an activity (a.k.a.~an event class). 
Moreover, a trace may have static and dynamic attributes. The former, \textit{trace attributes}, pertain to the whole trace and are shared by all events within that same trace. The value of a trace attribute remains the same throughout trace execution, i.e., it remains static. The latter (\emph{event attribute}) include, instead, in addition to the timestamp indicating the time in which the event has occurred, attributes such as the resource(s) involved in the execution of the corresponding activity, or other data specific to the event. In our motivating example in Section~\ref{sec:motivating}, 
the trace attributes are Bob's \emph{Application Type},  \emph{Loan Goal},  \emph{Requested Amount} and Bob's \emph{Credit Score}. An event attribute is instead the timestamp related to the time in which the activities in the trace have been carried.

\textbf{Definition 1 (Event).} \emph{An event is a tuple $(a, c, \text{time}, (d_1, v_1), \ldots, (d_m, v_m))$ where $a \in \Sigma$ is the activity name, $c$ is the case identifier, \textit{time} refers to the timestamp, and $(d_1, v_1), \ldots, (d_m, v_m)$ (with $m \geq 1$) are the event attributes and their values.}

\textbf{Definition 2 (Case).} \emph{A case is a non-empty sequence $\sigma = \langle e_1, \ldots, e_n \rangle$ of events such that $\forall i \in \{1, \ldots, n\}, e_i \in \mathbb{E}$ and $\forall i, j \in \{1, \ldots, n\}, e_i.c = e_j.c$, that is, all events in the sequence refer to the same case.}

We denote with $\Sigma$ the set of all the activity names, and with $\mathbb{E}$ the universe of all
events. A case is the sequence of events generated by a given process execution.

\textbf{Definition 3 (Trace).} \emph{A trace of a case $\sigma = \langle e_1, e_2, \ldots, e_n \rangle$ is the sequence of activity names in $\sigma$, $\langle e_1.a, e_2.a, \ldots, e_n.a \rangle$.}

We use $S$ to show all possible traces, and we use $sigma$ to show both cases and traces when there is no chance of confusion. The event log $\mathcal{L}$ is a collection of complete traces (i.e., the tracks documenting the completion of a complete process execution). For instance, we consider an event log that comprises a singular trace $\sigma_1$ in \figurename~\ref{fig:motivating_example}, pertaining to Bob's application process. The activity name of the first event in the case $\sigma_1$ is \act{Create Application} ($a$); this event occurred on the 27th of January ($time$) and refers to Bob's case ($c$).

Given a case $\sigma = \langle e_1, \ldots, e_n \rangle$ and a positive integer $k < n$, $\sigma_k = \langle e_1, \ldots, e_k \rangle$ is the prefix of $\sigma$ of length $k$. Furthermore, we define the prefix log as the log composed of all possible case prefixes, as is typically used in Predictive Process Monitoring settings~\cite{pmhandbook}. 
In \figurename~\ref{sec:motivating}, $\sigma_1$ refers to a prefix of Bob's case, where $k = 5$.

\textbf{Definition 4 (Prefix Log).} \emph{Given a log $\mathcal{L}$, the prefix log $\mathcal{L}^*$ of $\mathcal{L}$ is the event log that contains all prefixes of $\mathcal{L}$, i.e., $\mathcal{L^*} = \{\sigma_k : \sigma \in \mathcal{L}, 1 \leq k < |\sigma|\}$.}

\subsection{Declare}
\label{sec:declare}
When generating counterfactuals for a process execution, it is critical to determine whether these alternative scenarios preserve specific process 
constraints (expressed in the form of temporal constraints). 
The temporal constraints used in this paper are based on \textsc{Declare}, a language for describing declarative process models first introduced
in~\cite{declare}. A \textsc{Declare} model consists of a set of constraints applied to
(atomic) activities. 
These constraints, in essence, are built upon templates. Templates serve as abstract, parameterized representations of common patterns or structures inherent in the behaviour of the underlying process.
Templates have formal semantics expressed in Linear Temporal Logic for finite traces ($LTL_f$), making them verifiable and executable.
Table~\ref{table:declare} summarizes the main~\textsc{Declare} constructs used in this paper. The reader can refer to \cite{declare} for a full description of the language. 

For binary constraints (i.e., constraints specifying a relationship between two activities), one of the two activities is called \textit{activation}, and the other \textit{target}. When testing a trace for conformance over one of these constraints, the presence of the activation in the trace triggers
the clause verification, requiring the (non-)execution of an event containing the target in the
same trace. When the activation does not occur, the constraint is said to be \textit{vacuously satisfied}.  For example, 
the template $\Resp{\paramx}{\paramy}$ in Table~\ref{table:declare} is a binary template specifying that, should \activity{A} occur, then \activity{B} should occur after \activity{A}. In this case, \activity{A} would represent the activation, while \activity{B} the target. 
Unary templates (i.e., templates that refer only to 
a single activity), instead, define the cardinality or the position of the occurrence of an event
in a process instance. In the case of unary templates, we always consider the activities in the \declare constraints to be the activations. For instance, $\Exi{1}{\paramx}$ in Table~\ref{table:declare} is a unary template specifying that the activity \activity{A} should occur at least once in a process execution. In this case, \activity{A} represents the activation.

\begin{table}
\centering
\begin{tabular}{l p{90mm}}
\toprule
\textbf{Template} & \textbf{Description} \\
\midrule
$\Exi{1}{\paramx}$ & $\paramx$ occurs at \textbf{least} once\\
$\Abse{2}{\paramx}$ & $\paramx$ occurs at \textbf{most} once\\
$\ResEx{\paramx}{\paramy}$  & If $\paramx$ occurs, then $
\paramy$ occurs \\
$\Resp{\paramx}{\paramy}$ &
If {\paramx} occurs, then {\paramy} occurs after {\paramx} \\
$\AltRespShort{\paramx}{\paramy}$  & Each time {\paramx} occurs, then {\paramy} occurs after, before $\paramx$ occurs \\
$\ChaRes{\paramx}{\paramy}$ &  Each time {\paramx} occurs, then {\paramy} occurs immediately after \\
$\Prec{\paramx}{\paramy}$ &  $\paramy$ occurs only if preceded by $\paramx$ \\
\bottomrule
\end{tabular}
\caption{Main \textsc{Declare} templates}
\label{table:declare}
\end{table}

Given a \declare model and an event log, conformance checking is used to measure the conformance of the traces with respect to the model.
Rule checking is one of the techniques typically used to perform conformance checking of models described in terms of rules and constraints~\cite{pmhandbook}. The idea is to evaluate the conformance of the log with respect to rules, that is, constraints imposed by the process model. 
In the case of \declare, conformance checking measures how many (and which)
\declare constraints are satisfied or violated for a given event log. This is needed to measure the \emph{fitness} of the event log, i.e., how conformant the event log is regarding the \declare model~\cite{pmhandbook}. The more conformant an event log is, the higher its fitness score will be.
The formula for the rule-based log fitness\footnote{ From here on we use the term \textit{log fitness} for denoting the rule-based log fitness.} is hence defined as:
\[
\mathcal{F}_{\mathcal{L}}(\mathcal{M}_{decl}) = \frac{| \{ \varphi \in \mathcal{M}_{decl} \,|\,\forall \sigma \in \mathcal{L}, \sigma \models \varphi\}|}{| \mathcal{M}_{decl} |} 
\] where $\mathcal{L}$ represents the event log and $\mathcal{M}_{decl}$ the \declare model containing only the single \declare constraint $\varphi$.

In our case, we extend the definition of rule-based fitness from~\cite{pmhandbook}, to the case of a single trace, thus defining the notion of rule-based trace fitness \footnote{From here on we use the term \textit{trace fitness} for denoting the rule-based trace fitness.}: 
\[
\mathcal{F}_{\sigma}(\mathcal{M}_{decl}) = \frac{| \{ \varphi \in \mathcal{M}_{decl} \,|\,\sigma \models \varphi\}|}{| \mathcal{M}_{decl} |} 
\]
where $\sigma$ represents a single trace and $\mathcal{M}_{decl}$ the \declare model containing only the single \declare constraint $\varphi$.

For instance, given the trace $\sigma_1=\langle$\activity{a, b, c, a, b, c, d, a, b}$\rangle$ and a \textsc{Declare} model composed of two constraints $\Ini{\activity{a}}$ and $\ChaRes{\activity{a}}{\activity{b}}$, 
both constraints are fully satisfied as the trace starts with event \activity{a}, and every time \activity{a} happens, \activity{b} occurs right after (each of the three activations is satisfied). 
Since both the constraints of the model are satisfied, the trace fitness for $\sigma_1$ is $1$.
Given instead a trace $\sigma_2=$
$\langle$\activity{a, b, c, a, b, c, d, a}$\rangle$, the constraint $\ChaRes{\activity{a}}{\activity{b}}$ would be 
violated, so that only one out of the two constraints of the model is fulfilled by the trace. 
The trace fitness for $\sigma_2$  would hence be $0.5$.

\subsection{Predictive (Business) Process Monitoring}
\label{ch:ppmback}
Predictive Process Monitoring (PPM) is a branch of Process Mining that aims to use predictive techniques, such as Machine Learning models, to predict the future of running cases of a business process by using historical executions of completed traces (cases) from an event log~\cite{chiarappm1,tax2017predictive}.
Within PPM, there are several prediction tasks that one could aim to solve: predicting whether a certain predicate will be fulfilled or not~\cite{chiarappm2}, predicting the next activity or the continuation of a running incomplete case~\cite{pasquadibisceglie2019using,tax2017predictive}, predicting the outcome of a specific case~\cite{teinemaa2019outcome,irene}, or predicting the time it takes to complete a task or the time remaining until the completion of a case~\cite{van2011time}. Such prediction tasks support users in a decision-making process by offering an overview of the future of trace executions. For example, predicting the outcome of a case could allow a company to decide if and how resources or time should be allocated to cases depending on their outcome.

Training a predictive model usually requires preprocessing traces in such a way that is understandable for these models. For this reason, in the PPM literature, different encoding strategies are deployed to include the different attributes present in a trace. A popular encoding method used is the \emph{index-based encoding}, where the order of events is preserved, as opposed to other encoding techniques~\cite{irene}.

To give an example, each trace (prefix) $\sigma^m_i= <e_{i_1}, ... , e_{i_m} >$, $i = 1 ... k$, has to be represented through a feature vector $F_i = < {g_{i_1}},{g_{i_2}}, ..., {g_{i_h}} >$. In this paper, the simple-index and simple-trace index encodings are used.
\begin{itemize}[label=$\bullet$]
\item In the simple-index encoding~\cite{DBLP:conf/bpm/LeontjevaCFDM15}, the focus is on the events and on the order in which they occur in the trace. Each feature corresponds to a position in the trace, and the possible values for each feature are the event classes. The resulting feature vector $F_i$, for a trace prefix $\sigma_i^m$ of length $m$, is
$F_i = <a_{i_1}, ... ,a_{i_2}, ... ,a_{i_m}>$, where $a_{i_j}$ is the event class of the event at position $j$. Given the motivating example in Section~\ref{sec:motivating}
, Bob's case would take the following vector form: \\ $F_{Bob} = < \act{Create application},\act{Submit Documents},\act{Receive missing info email},\\ \act{Receive Reminder}, \act{Provide missing info}>$.
\item The simple-trace index encoding~\cite{caisepaper} leverages the same dynamic information related to the sequence of events as the simple-index encoding, but it also includes static information related to data (i.e., the trace attributes) in its feature vector. The resulting feature vector $F_i$, for the trace prefix $\sigma_i^m$ of length $m$, is represented as
$F_i = <s_i^1 , \dots , s_i^u , a_{i_1}, a_{i_2}, ..., a_{i_m}>$, where each $s_i$ is a static feature — corresponding to a trace attribute. Given the motivating example in Section~\ref{sec:motivating}
, Bob's case would take the following vector form: $F_{Bob} = <\text{New credit}, \text{Not specified}, 15000.0, 540, \act{Create application},\act{Submit Documents}, \\ \act{Receive missing info email},\act{Receive Reminder},\act{Provide missing info}>$.
\end{itemize}

\textbf{Definition 5 (Sequence/trace encoder).} A sequence (or trace) encoder $e: \mathbb{S} \rightarrow \mathcal{D}_1 \times \ldots \times \mathcal{D}_m$ is a function that takes a (prefix) trace $\sigma_i$ and transforms it into a feature vector $x_i = e(\sigma_i)$ in the $p$-dimensional vector space $D_1 \times \ldots \times \mathcal{D}_m$, with $\mathcal{D}_j$, $1 \leq j \leq p$ being the domain of the $j$-th feature.

\textbf{Definition 6 (Feature vector decoder).} A trace decoder $dec: \mathcal{D}_1 \times \ldots \times \mathcal{D}_m \rightarrow S$ is a function that takes a feature vector $x_i \in \mathcal{D}_1 \times \ldots \times \mathcal{D}_m$ and decodes it into a (prefix) trace $\sigma_i$.

For both encodings, we use the notation $F_i[k]$ to denote the k-th feature of the vector $F_i$. For instance, in the case of the simple-index encoding, $F_i[k] = a_{i_k}$. Moreover, we denote with $S$ the set of numeric indexes of static features and with $CF$ the set of numeric indexes related to control-flow features, that is $F_i[k]$ is a static feature if $k \in S$ and a control-flow one (an event class name) if $k \in CF$.

Although more encodings can be derived from the index encoding family, the focus in this paper is on the control-flow of the traces, which allows one to better control the generation of the counterfactual explanations that satisfy the background knowledge. The presented approaches in this paper could easily be extended to this other types of encodings as well.

\subsection{Counterfactual explanations}
\label{sec:expl}

Counterfactual explanations belong to the family of XAI methods. However, compared to other types of XAI methods, such as feature attribution methods~\cite{stierle2021bringing}, counterfactuals do not attempt to explain the inner logic of a predictive model but instead offer an alternative
to the user to obtain a desired prediction~\cite{wachter2017counterfactual}.
When dealing with black-box models, indeed, the internal logic of a model $h_{\theta}$ mapping a sample $\textbf{x}$ to a label $y$ (also called class value) is unknown, or otherwise uninterpretable to humans.

A counterfactual $\textbf{c}$ of $\textbf{x}$ is a sample for which the prediction of the black box is different from the one of $\textbf{x}$ (i.e., $h_{\theta}(\textbf{c}) \neq h_{\theta}(\textbf{x})$). For example, in the case of binary classification, if $h_{\theta}(\textbf{x}) = y$, i.e.,  the sample is predicted to belong to class $y$, a counterfactual should flip the prediction such that $h_{\theta}(\textbf{c})= y'$.
A counterfactual explainer is a function $f_k$, where $k$ is the number of requested counterfactuals, such that, for a given sample of interest $\textbf{x}$, a black box model $h_{\theta}$, and the set $\mathcal{X}$ of known samples from the training set, returns a set $C = \{\mathbf{c_1}, \ldots, \mathbf{c_h}\}$ of counterfactuals (with $h \leq k$), i.e., $f_k(\textbf{x}, h_\theta, \mathcal{X}) = C$.

\subsection{Genetic algorithms}
Genetic Algorithms (GAs) 
represent a powerful class of optimisation techniques, drawing inspiration from the natural processes of evolution. They have found widespread use across diverse domains due to their effectiveness in addressing intricate optimisation problems~\cite{geneticbook}.

GAs are fundamentally based on several key principles, acting as the pillars of their functioning. GAs operate within a population of potential solutions, evaluating their quality through a \textit{fitness function}. Building on the concept of ``survival of the fittest'', the algorithm selects the most promising solutions to serve as parents for the next generation based on the results of the \textit{fitness function}. These parent solutions undergo a set of genetic operations, 
where their genetic information is combined to create offsprings~\cite{geneticbook}. This new generation competes and evolves over successive iterations until one or more optimal solutions are identified. The GA terminates when the improvement with respect to the last generation is below a certain threshold, or when the maximum number of iterations is reached.

Each candidate solution in the search space, that is, each individual of the population, is described through a set of genes representing its \textit{chromosome or genotype}.
Besides the definition of the \textit{genotype representation} of a candidate solution, a GA usually requires the specification of the \textit{fitness function}, of the \textit{initial population},  as well as of the GA operators (\textit{selection}, \textit{crossover}, and \textit{mutation}).

The \textit{fitness function} defines the aspect(s) to optimise, thus helping select the best solution according to such aspect(s)~\cite{geneticbook}. Based on the nature of the optimisation problem, the aspect(s) to be optimised can be expressed in terms of a single objective or in terms of multiple objectives. 

In single-objective optimisation, the goal is to optimise a single, well-defined criterion. The fitness function, in this case, quantifies how well a solution satisfies the objective. It provides a clear measure of success, guiding the algorithm toward solutions that perform better with respect to the single criterion. Multi-objective optimisation, on the other hand, deals with problems where there are multiple, often conflicting, objectives to be considered simultaneously. In such cases, the fitness function becomes more complex as it needs to account for the trade-offs between different objectives. The challenge is to find a set of solutions that represent a trade-off between conflicting objectives, forming a Pareto front. The fitness function in multi-objective optimisation assesses how well a solution balances conflicting criteria. The choice between single and multi-objective optimisation depends on the problem at hand. 

The \textit{initial population} represents an initial set of candidate solutions. Usually, it is randomly generated; however, it can also be a set of already ``good'' solutions.

The \textit{crossover} operator determines how two parents share their genetic information. The quality of the parents significantly influences the quality of the offspring. This step helps to eliminate less desirable traits within the population. However, the offspring may inherit unwanted characteristics from the parents. The \textit{mutation} operator introduces random changes to the offspring's genetic makeup. These changes are determined by a mutation rate, representing the probability of altering specific traits within a given population. Mutation's primary role is to maintain diversity within the population. The \textit{selection} operator determines the members of the population that will be selected to be mated using the \textit{crossover} and \textit{mutaiton} operations. Different selection operators exist, such as random selection or tournament selection. In random selection, members of the population are chosen randomly for mating, while tournament selection picks random pairs of solutions from the population at each round, choosing the one that has the better \textit{fitness function} score~\cite{geneticbook}.

In the case of counterfactual generation, the initial population is a starting set of counterfactual candidates~\cite{geco,moc}. The \textit{fitness function} helps select the most promising counterfactuals to move into the mating phase, based on some desired properties. Depending upon the desired properties of the generated counterfactuals and possible trade-offs, the problem could be formulated as either a single-objective optimisation problem or a multi-objective problem. 
The fittest counterfactual solutions will undergo the crossover and mutations operators to produce better offsprings, i.e., more suitable counterfactual explanations given the input sample of interest~\cite{moc}. Both these operators help maintain the diversity of the set of generated counterfactuals $C$. 

\section{Related Work}\label{sec:relwork}
In this section, we will offer a detailed picture of the works related to
counterfactual explanations in general settings in Section~\ref{ch:cfs} and counterfactual explanation techniques specifically applied to Predictive Process Monitoring in Section~\ref{ch:ppmcf}

\subsection{Counterfactual explanations}
\label{ch:cfs}

Recently, Deep Learning has continued to outperform traditional statistical ML models across different domains. Despite this, in domains where critical decisions are made, such as the medical or the banking domain, such complex models have been generally avoided due to their inability to provide any reasoning for the prediction offered by such models.
To overcome the black-box issue, Explainable Artificial Intelligence (XAI), and in particular, counterfactual explanations, have emerged as a potential solution to offer a more in-depth look at how ML models and especially Deep Learning models function and what do they consider relevant from the input they receive.

Counterfactual explanations identify the smallest change in the input instance needed to flip the predicted outcome of a data instance~\cite{wachter2017counterfactual,dice}.
Counterfactual generation techniques can be classified in two main categories~\cite{verma2020counterfactual}: case-based and generative methods. We define case-based methods as methods that search for counterfactuals within the sample population by using a distance metric. Generative methods, instead, aim at generating synthetic counterfactuals that do not exist within the sample population through the use of an optimisation function that updates the input until the desired outcome is reached~\cite{dice,guidotti2018local}, or through Reinforcement Learning techniques, where an agent learns a reward function that changes the predicted outcome for a given sample~\cite{10.1145/3511808.3557429}. 
GAs are a popular optimisation-based solution used to create a population of potential counterfactual candidates by employing a fitness function~\cite{moc,created,geco}. In the literature, both single-objective and multi-objective GA solutions are presented for generating counterfactuals. While single-objective solutions may converge faster due to a lower 
complexity~\cite{geco}, multi-objective GAs have the advantage of providing a Pareto Front of optimal solutions, where trade-offs are made between different optimisation objectives~\cite{moc}.
The advantage of GAs is that they maximise the diversity of the population, whilst having the disadvantage that they are often stochastic, making it difficult to obtain consistent results.

One example of a single-objective solution for counterfactual generation is presented by~\citet{geco}, named Genetic Counterfactuals (GeCO). GeCO employs a single-objective genetic algorithm to fetch various counterfactuals, incorporating considerations of plausibility and accountability during mutation and crossover operations by utilizing a plausibility-feasibility constraint language outlined in the paper, named PLAF. These PLAF constraints only regard numerical features and aim to ensure that causal relationships between variables have to be maintained when generating counterfactuals. The downside of this approach is that these constraints have to be defined by a domain expert and cannot be extracted from data.

An example of a multi-objective genetic algorithm for the generation of counterfactuals is proposed by~\citet{moc}. The concept behind the Multi-objective Counterfactuals (MoC) involves conceptualizing the exploration of counterfactual instances as a multi-objective genetic optimisation problem. The resulting set of counterfactual instances, encompasses a diverse array of solutions that balances the various objectives. This balance is essential in achieving distinct trade-offs among the defined objectives. The choice of a multi-objective optimisation problem allows one to avoid the process of finding the optimal balance of weights between the different terms of a single-objective collapsed fitness function~\cite{moc}.

In many of the presented works, the feasibility of counterfactual explanations is presented as an important desideratum. Feasibility, or sometimes also called plausibility, should ensure that the counterfactuals that one generates are valid. This is done either by restricting the data manifold from where one could generate the counterfactuals~\cite{datamanifoldcfs}, specifying some constraints on individual or pairs of attributes, or minimizing the distance with respect to the closest point in the training set, when available.

However, no methods thus far leverage 
background knowledge to ensure that the generated solutions respect it. 
As mentioned by~\citet{priorxai} in their review of formal methods in XAI, no 
background knowledge injection has been explored for the generation of counterfactual explanations. As pointed out in their review on evolutionary algorithms (EAs) for XAI,~\citet{Zhou2023EvolutionaryAT} mention that one of the main challenges for EAs, and in XAI, is the lack of methods able to 
ensure that the generated explanations are in line with the background knowledge.

\subsection{Counterfactual explanations for Predictive Process Monitoring}
\label{ch:ppmcf}

The adoption of Deep Learning models in Predictive Process Monitoring (PPM) has synchronously brought upon the adoption of explanatory techniques intending to provide explanations for different prediction tasks~\cite{pmhandbook}. The challenge that arises with XAI in PPM is the way in which the data is structured. 
In static data, such as tabular data or images, every input has the same size, and the expectation is that all features will be present. In Process Mining, however, the format of the data is that of sequences that can be of varying lengths, thus resulting in a very dynamic setting, making the process of generating explanations much more complex.

Counterfactual explanations help users understand rather what to change to reach a desired outcome through what-if scenarios~\cite{verma2020counterfactual}. Four works so far have tackled the counterfactual explanation problem in the PPM domain~\cite{loreley,dice4el,created,caisepaper}. 

The first paper introduces LORELEY, an adaptation of the Local Rule-Based Explanations (LORE) framework~\cite{guidotti2018local}, which generates counterfactual explanations leveraging a surrogate decision tree model using a genetically generated neighbourhood of artificial data instances to be trained~\cite{loreley,guidotti2018local}. 
The prediction task the authors address is the one of multi-class outcome prediction. To ensure the generation of feasible counterfactuals, LORELEY imposes process constraints in the counterfactual generation process by using the whole prefix of activities as a single feature, encoding the whole control-flow execution as a variant of the process.

The second work presents \textsc{dice} for Event Logs (DICE4EL)~\cite{dice4el}, which extends a gradient-based optimisation method found in \textsc{dice} by adding a feasibility term to ensure that the generated counterfactuals maximize the likelihood of belonging to the training set. The prediction task addressed in the paper is that of next activity prediction with a milestone-aware focus.

The third, the most recent approach for generating counterfactual explanations for PPM, CREATED, leverages a genetic algorithm to generate candidate counterfactual sequences~\cite{created}. To ensure the feasibility of the data, the authors build a Markov Chain, where each event is a state. Then, using the transition probabilities from one state to another, they can determine how likely a counterfactual is, given the product of the states.

The fourth and final paper identified the need for a consolidated evaluation framework for assessing the efficacy of various counterfactual generation methods in PPM~\cite{caisepaper}.
The work aims to understand the differences in these techniques' performance and offer a standardized evaluation framework for counterfactual explanations in PPM. 
State-of-the-art metrics are adapted, and a novel evaluation metrics is introduced for measuring the compliance of counterfactuals to some process 
background knowledge, described in terms of 
\declare constraints. The obtained results 
highlight the need for
techniques able to meet 
process-specific desiderata for trace counterfactuals.

It can be noted that none of the previous approaches make use of background knowledge explicitly when generating counterfactual explanations. However, as mentioned in Section~\ref{ch:cfs}, background knowledge can play an important role in ensuring the feasibility of the generated counterfactuals, especially from the control-flow perspective, where different constraints may have a different impact on the outcome of a trace execution. Furthermore, none of the previous approaches have posed the counterfactual generation problem for Predictive Process Monitoring as a multi-objective optimisation problem. The present work aims to specifically fill these two gaps identified in the literature.

\section{Methodology}\label{sec:approach}
In this section, we first focus on the general counterfactual explanations (Section~\ref{sec:c_general}). In detail, 
we first describe the desiderata that general counterfactual explanations should satisfy. We then contribute to the literature by providing a novel mapping from desiderata to concrete optimisation objectives for optimisation-based counterfactual generation techniques. This  addresses a gap in the current body of knowledge, as prior research has not systematically explored the alignment of desired counterfactual qualities with optimisation goals. We then describe two existing solutions for the generation of counterfactual explanations through genetic algorithms (GAs). 

Afterwards, in Section~\ref{sec:c_PPM}, we focus on the PPM scenario, by introducing two methodologies for the counterfactual generation process through GAs when dealing with counterfactuals for PPM. 
In detail, we first introduce the adaptations required in terms of desiderata (Section~\ref{sec:desiderata_ppm}), we then present how these adapted desiderata can be satisfied through the adaptation of the single and multi-objective fitness functions (Section~\ref{sec:problemformulation}), and we finally introduce the updates carried out to the crossover and mutation operators of the proposed GAs for the generation of counterfactuals for PPM (Section~\ref{sec:adaptedoperators})

\subsection{From counterfactual desiderata to GA objectives}

\label{sec:c_general}
In this section, we report the desiderata for counterfactual explanations presented in~\cite{verma2020counterfactual}.

These counterfactual explanations desiderata can be summarised in the following six categories:
\begin{enumerate}[label={\textbf{\act{(D\arabic*)}}}]
    \item \textbf{Validity:} Counterfactual explanations should change features in a way that the prediction of the original input is flipped, thus ensuring that the explanation aligns with the desired class.
    \item \textbf{Input Closeness:} Effective counterfactuals should minimize the change to provide a more understandable explanation.
    \item \textbf{Sparsity:} Effective counterfactuals should change as few attributes as possible to provide a more concise explanation.
    \item \textbf{Plausibility:} Counterfactuals must be plausible by adhering to observed correlations among features in the training data, thus ensuring they are feasible and realistic.
    \item \textbf{Causality:} Acknowledging the interdependence of features, counterfactuals should respect known causal relations between features to be both realistic and actionable.
    \item \textbf{Diversity:} When generating a set of counterfactuals, the differences among the counterfactuals in the set should be maximised 
    to provide different alternatives that a user could choose from.
\end{enumerate}

When considering optimisation-based techniques for generating counterfactual explanations, these desiderata can be targeted by  directly translating them into optimisation objectives. In their work,~\citet{geco} and~\citet{moc} instantiate only the first four desiderata into different optimisation objectives for the fitness function.

The validity of a counterfactual \( \mathbf{c} \)  (desideratum \act{D1}) is translated into an objective $o_1$ that is measured as the difference in probability between \( \mathbf{b}(\mathbf{c}) \) and \( y' \). It is defined as:
\begin{align*}
o_1(h_{\theta}(\mathbf{c}), y') = \begin{cases}
    0 & \text{if } h_{\theta}(\mathbf{c}) = y', \\
    \inf_{y} |h_{\theta}(\mathbf{c}) - y'| & \text{otherwise.}
\end{cases}
\end{align*}
where $\inf_{y}$ refers to infimum, or lowest greater bound over the possible values of $y$, measuring the closeness between the predicted value $h_{\theta}(\mathbf{c})$ and the target $y'$.

Input closeness of the counterfactual \( \mathbf{c} \) to \( \mathbf{x} \) (desideratum \act{D2}) is targeted through an objective $o_2$ that 
measures the distance between the two values in terms of 
the edit distance for categorical variables, or a numerical distance between the continuous variables. It is defined as:
\begin{align*}
o_2(\mathbf{x}, \mathbf{c}) &= \frac{1}{f} \sum_{j=1}^{f} d(x_j, c_j)
&
d(x_j, c_j) &= \begin{cases}
    ||x_j - c_j||_1 & \text{if } x_j \text{ is numerical}, \\
    I_{x_j \neq c_j} & \text{if } x_j \text{ is categorical},
\end{cases}
\end{align*}
where \( f \) is the number of features, and \( d(x_j, c_j) \) depends on the feature type. When the feature is numerical, the \( L_1 \) norm is used, while if the feature is categorical, the indicator function is used to check whether $c_j$ matches $x_j$.

The sparsity of the counterfactual \( \mathbf{c} \) with respect to \( \mathbf{x} \) (desideratum \act{D3}) is translated into an objective $o_3$ that measures the number of changes in the counterfactual in terms of 

the \( L_0 \) norm:
\begin{equation}
o_3(\mathbf{x}, \mathbf{c}) = ||\mathbf{x} - \mathbf{c}||_0 = \sum_{j=1}^{f} I_{x_j \neq c_j}.
\end{equation}

The plausibility of a counterfactual \( \mathbf{c} \) (desideratum \act{D4}) is targeted through an object $o_4$, measured in terms of the distance between the counterfactual  \( \mathbf{c} \) and the closest point in the reference population $X$. This measure allows for determining the implausibility of the candidate in the context of the reference population: 
\begin{equation}
    o_4(\mathbf{c},\mathcal{X}) =  \min_{\mathbf{x} \in \mathcal{X}} d(\mathbf{c},\mathbf{x})
\end{equation}

The fifth desideratum (\act{D5}), causality, is usually addressed through the insertion of constraints during the GA's optimisation process. As mentioned in Section~\ref{ch:cfs},~\citet{geco} make use of PLAF constraints to ensure that the generated counterfactuals satisfy causal relationships between features. However, these PLAF constraints only regard numerical features and cannot account for temporal relationships between categorical attributes.

When using GAs, the diversity desideratum, (\act{D6}), is addressed by construction, as the main advantage of GAs is the maximisation of the diversity of the population, due to crossover and mutation operators that enhance the diversity.

We incorporate two optimisation methods grounded in genetic algorithms~\cite{geco,moc}, strategically positioned as essential benchmarks to establish a comparative baseline for the framework under consideration. By introducing these optimisation methods, we aim to contextualize and evaluate the performance of our proposed framework against established techniques, thus highlighting its innovation and results. Therefore, we denote the two optimisation methods as baselines in our study.

The resulting fitness function, considering the four desiderata addressed by the four objectives $o_1, \ldots, o_4$ described above, 
results in the following 
single-objective fitness function for the GA, which we call $\mathbf{f_{baseline}^{single}}$:
\begin{equation}
\mathbf{f_{baseline}^{single}} :=   o_1(h_{\theta}(\mathbf{c}),y) + \alpha \cdot o_2(\mathbf{x},\mathbf{c}) + \beta \cdot o_3(\mathbf{x},\mathbf{c}) + \gamma \cdot o_4(\mathbf{c},\mathcal{X}) 
\label{eq:sobjbase}
\end{equation}
\begin{align*}
\text{where } \alpha, \beta, \gamma,
& \text{ are weighting factors for each term, controlling their influence on the fitness.}
\end{align*}

A natural solution to the complex problem of balancing the different objectives posed in the previous single-objective problem is moving to a multi-objective problem. In the multi-objective setting, a fitness function is constructed to evaluate solutions based on multiple objectives, allowing for a more nuanced representation of trade-offs. In their work,~\citet{moc} propose a fitness function based on the same four desiderata, foregoing the causality desideratum \act{D5}, and addressing the diversity desideratum \act{D6} by employing a multi-objective optimisation technique. The 
multi-objective fitness function, which we call $\mathbf{F_{baseline}^{multi}}$, takes the following form:

\begin{equation}
\mathbf{F_{baseline}^{multi}} :=  (o_1(h_{\theta}(\mathbf{c}),y), o_2(\mathbf{x},\mathbf{c}), o_3(\mathbf{x},\mathbf{c}), o_4(\mathbf{c},\mathcal{X}))
  \label{eq:mobjbase}
\end{equation}
where the different objectives are the same as the ones detailed above.
In this case, the objectives are no longer weighted and collapsed into a single-objective function. A trade-off among the different objectives can rather be
directly extracted from the Pareto Set of the solutions, thus also reinforcing the diversity between the different counterfactuals.

\subsection{Counterfactuals explanation for PPM with GAs}
\label{sec:c_PPM}
In this section, we introduce the proposed approaches for the generation of counterfactual explanations, starting from the adaptation of the generic desiderata and hence GAs approaches to the domain of Predictive Process Monitoring. We start by describing the adaptations in terms of desiderata, and we then report how these adaptations are reflected in the formulation of the fitness functions (both single-objective and multi-objective), as well as in terms of the mating operators.

\subsubsection{Adaptation of desiderata for PPM}
\label{sec:desiderata_ppm}
When creating counterfactual explanations for Predictive Process Monitoring, the required desiderata can mostly be directly taken from 
the general case. However, in the context of process mining, process relationships, as control flow relationships among events, have to be also preserved. A counterfactual, indeed, should not violate control flow relationships among activities. For instance, a counterfactual cannot pretend that a loan application submission is performed before a loan application is filled, as in the case of Bob (see Section~\ref{sec:motivating}) or that a loan request is approved before it is evaluated.

The optimisation function should hence consider: (1) counterfactual validity (\act{D1}), (2) similarity to the query instance (\act{D2}); (3) sparsity in the feature vector (\act{D3});
(4) similarity to the training data (\act{D4}); (5) adherence to some knowledge making explicit the temporal ordering relationship among activities (\bk) (\act{D5}).

Having translated these desiderata to the PPM domain, we now move on to the formulation of the optimisation problem for the generation of counterfactual explanations in the  field of PPM. We use the baseline previously introduced as a starting point and expand the objective function for both the single and multi-objective case.

\subsubsection{Optimisation problem formulations for counterfactual explanations in PPM}
\label{sec:problemformulation}
In the previous section, we introduced 
the formulation of the optimisation problem for the generation of general counterfactual explanations in both the single-objective and multi-objective scenarios. 

To customize the general formulation to the case of Predictive Process Monitoring, we need to consider the updated desiderata. To this aim we update the optimisation functions so that they consider also the desideratum \act{D5}.

To extend the baseline fitness function previously introduced, a fifth objective \( o_5 \) is introduced, aiming to address the causality desideratum directly \act{D5} in the Predictive Process Monitoring context, that is the adherence to \bk on the control flow, that we assume to be available in the form of \textsc{Declare} constraints. To measure this aspect, this objective leverages the trace fitness metrics defined in Section~\ref{sec:declare}, which measures the number of satisfied temporal constraints present in the \textsc{Declare} model $\mathcal{M}_{Decl}$, leveraged as \bk related to the process underlying the data. This component is  
 a way to check how fit \( \mathbf{c} \) is with respect to \( \mathbf{x} \):

\begin{equation}
o_5(\mathbf{x}, \mathbf{c}, \mathcal{M}_{Decl}) = F_{dec(\mathbf{x})}(\mathcal{M}_{Decl}) \equiv F_{dec(\mathbf{c})}(\mathcal{M}_{Decl}))
\end{equation}
where, for each \textsc{Declare} constraint $\varphi \in \mathcal{M}_{Decl}$, if \( dec(\mathbf{x}) \models \varphi\), that is, if $d(\textbf{x})$ satisfies the proposition $\varphi$, then $dec(\mathbf{c})$ should also satisfy the proposition $\varphi$. The $dec$ function represents the decoder function, allowing us to go from the feature vector representation to the trace representation.

The adapted single-objective fitness function, which we call $\mathbf{f_{adapted}^{single}}$, takes the following form:
\begin{equation}
\mathbf{f_{adapted}^{single}} :=   \begin{small} o_1(h_{\theta}(\mathbf{c}),y) + \alpha \cdot o_2(\mathbf{x},\mathbf{c}) + \beta \cdot o_3(\mathbf{x},\mathbf{c}) + \begin{small}\gamma\end{small} \cdot o_4(\mathbf{x},\mathbf{c},X_{obs}) + \delta \cdot o_5(\mathbf{x},\mathbf{c},\mathcal{M}_{Decl})\end{small}
\label{eq:sobjadapted}
\end{equation}
where
$\alpha, \beta, \gamma, \delta$ are weighting factors for each term, controlling their influence on the fitness.

However, in this setting,
if one wants to obtain counterfactuals close to the query instance, the difference between counterfactual outcome and
desired outcome becomes more difficult to minimize while minimizing also the distance to the input query instance and the conformance with respect to the \bk. For this reason, we also extend the counterfactual generation problem to a multi-objective optimisation setting.

In the multi-objective case, the fifth objective \( o_5 \) is once again introduced, to address the causality desideratum (\act{D5}) in the PPM context, i.e., the adherence to the \bk, 
assuming its availability in the form of \textsc{Declare} constraints. We highlight the fact that  multi-objective optimisation for counterfactual generation within the context of PPM has not been explored before. The multi-objective optimisation adapted fitness function, which we call $\mathbf{F_{adapted}^{multi}}$, takes the following form:

\begin{equation}
\mathbf{F_{adapted}^{multi}} :=  (o_1 (h_{\theta}(\mathbf{c}),y), o_2(\mathbf{x},\mathbf{c}), o_3(x,\mathbf{c}), o_4(\mathbf{c}, \mathcal{X}),o_5(\mathbf{x},\mathbf{c},\mathcal{M}_{Decl})
\label{eq:mobjadapted}
\end{equation}

where the different objectives are the same as the ones detailed for the single-objective function.

In this case, however, as for the general case, the objectives are no longer weighted and collapsed into a single-objective function. A trade-off among the different objectives can rather be 
directly extracted from the Pareto Set of the solutions, ensuring the diversity between the different counterfactuals.

\subsubsection{Temporal knowledge-aware crossover and mutation operators}
\label{sec:adaptedoperators}
In this section, we introduce both a novel crossover and a novel mutation operator that integrate~\bk, represented as \textsc{Declare} constraints.
These operators replace the baseline crossover and mutation operators in  the adapted genetic algorithms -- both in the single-objective and in the multi-objective case.

The idea behind the adapted crossover and mutation operators is generating offspring individuals that preserve the satisfaction\footnote{We consider as satisfied only non-vacuously satisfied constraints.} of the \declare constraints of the original query, while still encouraging the diversity within the generated offspring. Intuitively, this means that the adapted operators should change the features of the original query instance as the classical operators, while being careful that (i) the constraints satisfied in the original query instance are not violated in the offspring individual (that is, the target activities in the constraints remain unchanged); (ii) the constraints that are not activated in the original query instance are not activated in the offspring individuals (that is, no new activation activities are introduced in the offspring instance). To this aim, starting from a \declare model $\mathcal{M}_{decl}$, we build the set of all the activation activities $A$ and the set of all the target activities $T$. Assuming that $\Sigma_{\mathcal{M}_{decl}}$ is the alphabet of the activities used for instantiating the \declare templates and generating the constraints in $\mathcal{M}_{decl}$, $A=\{a \in \Sigma_{\mathcal{M}_{decl}}|$ $\exists$ $\varphi \in \mathcal{M}_{decl}$ and $a$ is an activation of $\varphi\}$ and $T=\{a \in \Sigma_{\mathcal{M}_{decl}} |$ $\exists \varphi \in \mathcal{M}_{decl}$ and $a$ is a target of $\varphi\}$. 

\begin{figure}[ht]
\centering
  \includegraphics[width=\textwidth]{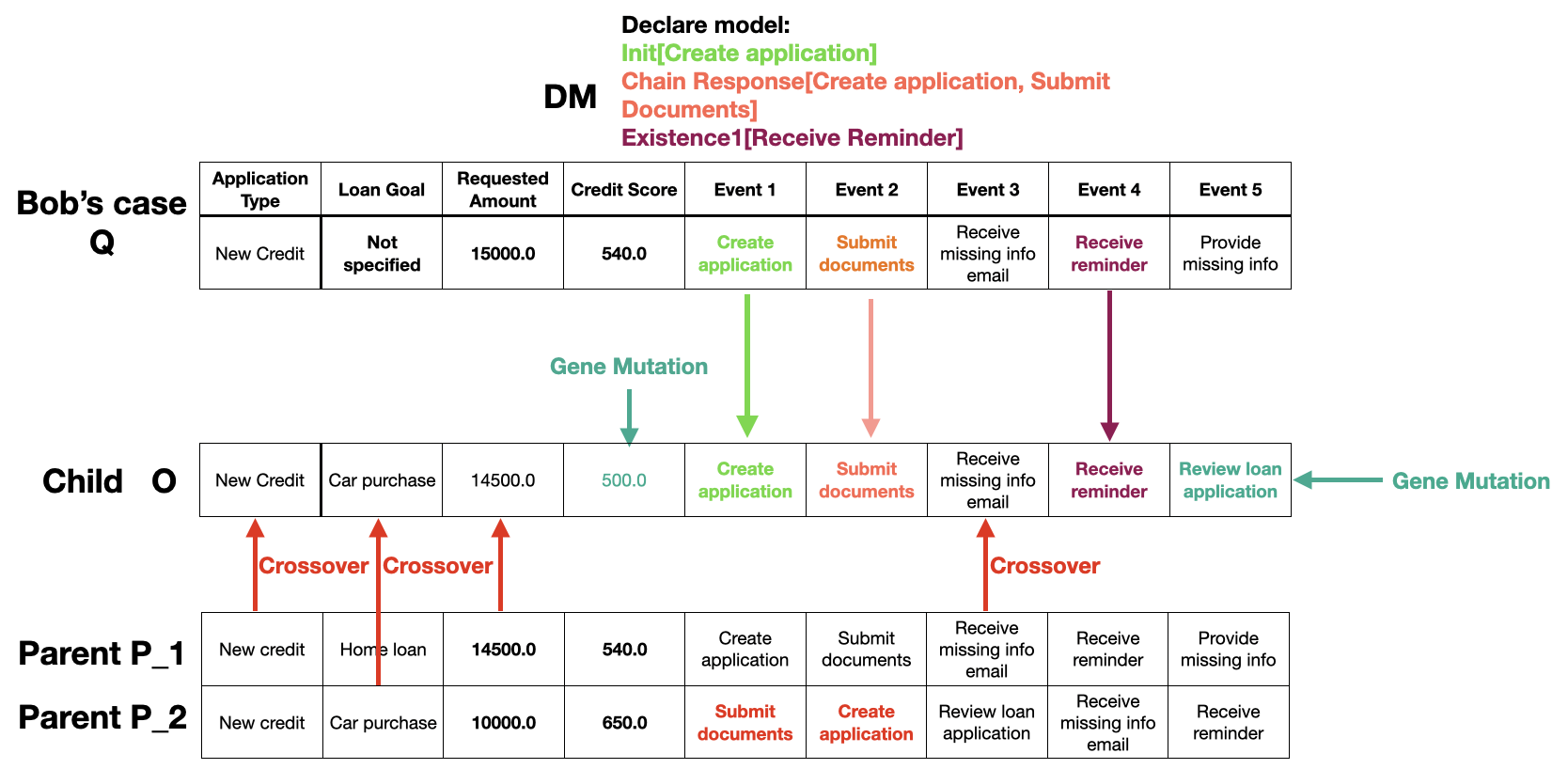}
  \caption{Example showcasing the functioning of the adapted genetic operator.}
  \label{fig:adapted}
\end{figure}

For instance, in the case of Bob 
from the motivating example in Section~\ref{sec:motivating}, we can assume to have a \declare model $\mathcal{M}_{decl}$(reported at the top of \figurename~\ref{fig:adapted}), provided to us by an expert, that represents our \bk.  
$\mathcal{M}_{decl}$ is composed of 3 \declare constraints: $\Ini{\act{Create application}}$,~$\ChaRes{\act{Create application}}{\act{Submit documents}}$,
~$\Abse{2}{\act{Receive reminder}}$. According to $\mathcal{M}_{Decl}$,  
every application has to begin with the \act{Create application} activity, that has to be directly followed by the \act{Submit documents} activity. Finally, the activity \act{Receive Reminder} can be performed at most once during the processing of a loan application.

Given $\mathcal{M}_{decl}$, the set of activations is $A =\{\act{Create application}, \act{Receive reminder}\}$ and the set of target activities is $T = \{\act{Submit documents}\}$.

\textbf{Crossover:}
The Adapted Crossover Operator, presented in Algorithm~\ref{alg:crossoveroperator}, is designed for generating offspring individuals from two parent individuals ($\mathbf{P_1}$ and $\mathbf{P_2}$), as the classical crossover operator, while however guaranteeing that the \declare constraints satisfied in the original query instance $\mathbf{Q}$ are still satisfied for the offspring $\mathbf{O}$. It takes as input the original query instance $\mathbf{Q}$, two parent individuals, $\mathbf{P_1}$ and $\mathbf{P_2}$, the crossover probability $p_c$, the activation, and target sets $T$ and $A$ extracted from the \declare constraints, as well as the indexes for the control-flow features $CF$.

The Adapted Crossover Operator starts by initiating an offspring individual ($\mathbf{O}$) by retaining from $\mathbf{Q}$ the control flow phenotype that enables the satisfaction of the \declare constraints, that is, any activities that appear in either the target set $T$ or activation set $A$ (lines 1-7). This ensures that any of the \declare constraints activated and satisfied in $\mathbf{Q}$ are also satisfied in $\mathbf{O}$. Next, for each feature, a random probability $p$ is sampled for the choice between the two parents $\mathbf{P_1}$ or $\mathbf{P_2}$ (line 8). The empty features in the offspring individual phenotype are then filled with one of the two parents' genetic material but only if the parent's control flow features are not activation activities, that is, they do not belong to the set $A$ (lines 8-18). Specifically, the crossover probability ($p_c$), together with the drawn probability $p$, guides the random selection of genetic material from either parent $\mathbf{P_1}$ or $\mathbf{P_2}$ as in classical crossover operators, if the selected item is not an activation activity. Otherwise, the crossover operator resorts to using the corresponding segment from the original query instance $\mathbf{Q}$. This ensures that no new constraint activations are introduced in the offspring population.

\begin{algorithm}
\caption{Adapted Crossover Operator}
\label{alg:crossoveroperator}
\begin{algorithmic}[1]
\Require Parent Individuals $\mathbf{P_1}$ and $\mathbf{P_2}$, Crossover Probability $p_c$, Original Query Instance $\mathbf{Q}$, activation $A$ and target $T$ activity sets extracted from the \declare constraints, 
control-flow indexes $CF$
\Ensure Offspring Individual $\mathbf{O}$
\For{$i$ from $1$ to $|\mathbf{Q}|$}
    \If{$\mathbf{Q}[i] \in T \cup A$}
        \State $\mathbf{O}[i]=\mathbf{Q}[i]$
    \Else
        \State $\mathbf{O}[i]=$null
    \EndIf
\EndFor
\For{$i$ from $1$ to $|\mathbf{O}|$}
    \State $p \sim U(0, 1)$
   \If{$\mathbf{O}[i]$ is null}
        \If{$p < p_c \wedge (i \notin CF \vee {\mathbf{P_1}}[i] \notin A$})
            \State $\mathbf{O}[i] = {\mathbf{P_1}}[i]$
        \ElsIf{$p > p_c \wedge (i \notin CF \vee {\mathbf{P_2}}[i] \notin A$})
            \State $\mathbf{O}[i] = {\mathbf{P_2}}[i]$
        \Else
            \State $\mathbf{O}[i]=\mathbf{Q}[i]$
        \EndIf
  \EndIf
\EndFor

\State \textbf{return} $\mathbf{O}$
\end{algorithmic}
\end{algorithm}

Figure~\ref{fig:adapted} showcases how the adapted crossover operator could work in the case of the motivating example of Bob introduced in Section~\ref{sec:motivating}. 
Given Bob's case, two parents from the previous population and the sets $A$ and $T$, the adapted crossover operator starts by checking the part of Bob's case that is present in either of the two sets, and then imposes that onto the offspring. Specifically, it can be seen that the values of the control-flow features 
\attr{Event 1}, \attr{Event 2} and \attr{Event 4} appear either in $T$ or in $A$,
thus the following activities \act{Create application}, \act{Submit documents}, and \act{Receive reminder} are directly imposed onto the offspring individual $\mathbf{O}$.

For the other event features, the gene \attr{Event 3} = \act{Receive missing info email} is crossed over from $\mathbf{P_1}$, while \attr{Event 5} = \act{Update missing info} is crossed over from $\mathbf{P_2}$. For the non-control-flow features, since they are attributes that are not related to the control-flow of the trace, the genes are randomly chosen from either $\mathbf{P_1}$ or $\mathbf{P_2}$.

\textbf{Adapted Mutation operator:}
The Adapted Mutation Operator, presented in Algorithm~\ref{alg:mutation_operator}, focuses on mutating an offspring individual ($\mathbf{O}$) while preserving the satisfaction of the temporal constraints. 
The operator takes as input the offspring $\mathbf{O}$, the mutation probability $p_m$, the set of the domain of the features $D=\bigcup{\mathcal{D}_i}$ extracted from the trace encoder, the set of the activation activities $A$, and the control-flow indexes $CF$. The operator returns as output a mutated offspring individual. By domain of the features, we refer to the possible values that a certain attribute can take.

The algorithm starts by sampling a random probability of mutation $p$  (line 1). The algorithm then iterates through each feature from 1 to $|O|$ (lines 2-9). For control-flow features, if the probability of mutation is under the set threshold probability $p_m$, the value of $O[i]$ is then uniformly sampled from the set $\mathcal{D}_{i} \setminus A$, i.e., excluding activities in the set $A$ from the domain of the i-th feature to avoid introducing new constraint activations and hence potential violations (lines 4-5). 
For non-control-flow features, if the mutation probability $p$ is lower than the predefined threshold $p_m$, the value of $O[i]$ is sampled uniformly from the set $\mathcal{D}_{i}$ (lines 6-7).
The algorithm proceeds mutating each feature accordingly, and the mutated offspring $\mathbf{O}$ is returned as the result. This operator ensures that mutations align with temporal constraints, contributing to the diversity and adaptability of the genetic algorithm.

\begin{algorithm}
\caption{Adapted Mutation Operator}
\label{alg:mutation_operator}
\begin{algorithmic}[1]
\Require Offspring $\mathbf{O}$, Mutation Probability $p_m$,  temporal constraint activations $A$, Set of the domain of the features 
$D$, control-flow features indexes $CF$
\Ensure Mutated Offspring $\mathbf{O}$
    \State $p \sim U(0, 1)$
\For{$i$ from $1$ to $|\mathbf{O}|$}
    \State $p \sim U(0, 1)$
    \If{$p < p_m \wedge i \in CF$}
        \State $\mathbf{O}[i] \sim U(\mathcal{D}_{i} \setminus A)$
    \ElsIf{$p < p_m \wedge i \notin CF$}
        \State $\mathbf{O}[i] \sim U(\mathcal{D}_{i})$

    \EndIf
\EndFor
\State \textbf{return} $\mathbf{O}$
\end{algorithmic}
\end{algorithm}

Sticking with Bob's case, the mutation operator then mutates the gene \attr{Event 5} using the adapted mutation operator, picking a value in the domain of the activities while ensuring that we exclude the activities in the activation set $A$, that is \act{Create application} and \act{Receive reminder}, from the set of possible values. In this way, we avoid a violation of the constraint $\ChaRes{\act{Create application}}{\act{Submit documents}}$, as well as of the constraint $\Abse{2}{\act{Receive reminder}}$, considering that \act{Receive reminder} has already been performed in \attr{Event 3}. The \attr{Credit Score} attribute, is also mutated, by randomly sampling from the distribution of its domain. 

\section{Evaluation}
\label{sec:eval}

To evaluate the contributions proposed for the generation of counterfactuals for PPM, we focus on answering the following research questions:

\begin{enumerate}[label={\textbf{(RQ\arabic*)}}]
    \item How do the adapted fitness functions and genetic operators impact the quality of the generated counterfactual explanations?
    \item What are the differences between single-objective optimisation and multi-objective for generating counterfactuals in Predictive Process Monitoring (PPM)?
\end{enumerate}

While the first research question (\textbf{RQ1}) focuses on the impact of the adapted fitness functions and the new crossover and mutation operators, the second (\textbf{RQ2}) aims at comparing the single and the multi-objective formulations of the problem for the counterfactual explanation in PPM.

To answer these research questions, our focus lies on evaluating and comparing various approaches. This involves leveraging either the \emph{baseline or adapted fitness function}—be it single-objective or multi-objective—of the optimisation problem, coupled with the \emph{baseline or adapted crossover and mutation operators}. This approach results in the identification of four distinct counterfactual generation strategies:
\begin{itemize}
    \item \boso (Baseline Operators/fitness function and Single-objective optimisation) using the baseline crossover \& mutation operators and the baseline fitness function in terms of single-objective optimisation as reported in Eq.~\ref{eq:sobjbase}~\cite{caisepaper}.
    \item \aoso (Adapted Operators/fitness function and Single-objective optimisation) using the adapted crossover \& mutation operators and the adapted fitness function in terms of single-objective optimisation found as reported in Eq.~\ref{eq:sobjadapted}.
    \item \bomo (Baseline Operators/fitness function and Multi-objective optimisation) using the baseline crossover \& mutation  operators and the baseline fitness function in terms of multi-objective optimisation as reported in Eq.~\ref{eq:mobjbase}~\cite{moc}. 
    \item \aomo (Adapted Operators/fitness function and Multi-objective optimisation) using adapted crossover \& mutation operators and the adapted fitness function in terms of multi-objective optimisation as reported in Eq.~\ref{eq:mobjadapted}. 
\end{itemize}

In the following sections we detail the datasets used for carrying out the comparison (Section~\ref{sec:dataset}), the metrics used for the evaluation (Section~\ref{sec:metrics}) and the experimental settings (Section~\ref{sec:setting}).

\subsection{Datasets}
\label{sec:dataset}

To compare the counterfactual approaches, we selected 5 different publicly available real-life event logs, 
labelled according to the satisfaction of their traces with respect to the $11$ \declare constraints shown in Table~\ref{tab:outcome}. We used, in total, $15$ datasets (combinations of event logs and labellings). Most of the datasets are taken from the Business Process Intelligence Challenge (BPIC), a challenge held every year where participants can apply various Process Mining techniques to uncover insights from the data. The labellings used in this paper are the same ones used by~\citet{irene}.

The event logs were chosen due to the wide range of different characteristics, such as the number of trace attributes, varying trace lengths and log size. 
Table~\ref{tab:datastat} reports, for each dataset, the number of traces, variants (i.e., unique sequences of event classes), event attributes, and trace attributes of the log,
as well as the constraint used for labelling the log and the range of prefix lengths  inspected for the counterfactual generation. 

\begin{table}[tb]
\centering
\begin{tabular}{l}
\scalebox{.9}
{\parbox{.5\linewidth}{%
\begin{alignat*}{2}
&\phi_1 &=& \Exi{1}{\act{Accept loan application}} \\
&\phi_2 &=& \Exi{1}{\act{Cancel Loan Application}} \\ 
&\phi_3 &=& \Exi{1}{\act{Reject Loan Application}} \\ 
&\phi_4 &=& \Resp{\act{Send confirmation receipt}}{\act{Retrieve missing data}} \\
&\phi_5 &=& \Exi{1}{\act{Accept loan application}} \\
&\phi_6 &=& \Exi{1}{\act{Cancel Loan Application}} \\
&\phi_7 &=& \Exi{1}{\act{Reject Loan Application}} \\ 
&\phi_8 &=& \Exi{1}{\act{{Patient returns to the emergency room within 28 days from  discharge}}} \\
& \phi_9 &=& \Exi{1}{\act{Patient is (eventually) admitted to intensive care}}  \\ 
&\phi_{10} &=& \Exi{1}{\act{Patient is discharged from the hospital on the basis of something other than Release A}} \\
&\phi_{11} &=& \Exi{1}{\act{Public Contest Recourse}}
\end{alignat*}}}
\end{tabular}
\caption{The \textsc{Declare} constraints used for the labellings}
\label{tab:outcome}
\end{table}

\textbf{BPIC2012.}~\cite{bpic2012}
This dataset, which was first made available in connection with the Business Process Intelligence Challenge (BPIC) in 2012, includes the loan application process execution history for a Dutch banking institution. Every instance in this log documents the events surrounding a certain loan application.
Labellings were established for classification reasons depending upon the outcome of a case, i.e., whether the application is approved, denied, or cancelled. This seems to be a multi-class classification problem intuitively. However, it is treated as three distinct binary classification tasks to maintain consistency with the binary task of outcome-oriented predictive process monitoring.
These tasks are called BPIC2012\_1, BPIC2012\_2, and BPIC2012\_3 in the experiments. From each event log, three separate labellings are created, namely $\phi_1$ for BPIC2012\_1, $\phi_2$ for BPIC2012\_2, and $\phi_3$ for BPIC2012\_3 reported in Table~\ref{tab:outcome}.

\textbf{BPIC2015.}~\cite{bpic2015}
This dataset consists of event logs from 5 different municipalities from the Netherlands, regarding a building permit application process. The data from each municipality is treated as a separate
event log and a single labelling function is applied to each one. The labelling function is based on the satisfaction/violation of the \declare constraint $\phi_4$ in Table~\ref{tab:outcome}. 
The prediction task for each of the 5 municipalities is denoted as BPIC2015\_i, where $i = 1 . . . 5$ indicates the number of the
municipality. 

\textbf{BPIC2017.}~\cite{bpic2017}
This event log originates from the same financial institution as the BPIC2012 one.
However, the data collection has been improved, resulting in a richer and cleaner dataset. As in the
previous case, the event log records execution traces of a loan application process. 

Similarly to BPIC2012, three separate labellings are created: $\phi_5$ for BPIC2017\_1, $\phi_6$ for  BPIC2017\_2, and $\phi_7$ for BPIC2017\_3 reported in Table~\ref{tab:outcome}.

\textbf{Sepsis.}~\cite{sepsis}
This event log records cases of patients with possible symptoms of a life-threatening sepsis condition in a hospital in the Netherlands. Each case belongs to a patient's admission into the emergency room up until their discharge from the hospital. Laboratory tests are recorded as events, and the discharge reason is available in the data in an obfuscated manner. From this event log, three separate labellings are created, namely $\phi_8$ for Sepsis\_1, $\phi_9$ for Sepsis\_2, and $\phi_{10}$ for Sepsis\_3, that can be found in Table~\ref{tab:outcome}.

\textbf{Legal Complaints.}~\cite{legalcomplaints}
This event log records execution traces of a public contract auction in Italy and whether any recourses have been registered. A public contract auction is when the Public Administration of a region requires a service to be done and holds a public contest where private contractors can bid to provide the service. Whoever wins the contest, is obligated to carry out the service. A single labelling is used to determine whether there is a recourse to the public contest or not, defined in Table~\ref{tab:outcome} as $\phi_{11}$.

\begin{table}[bt]
\centering
\scalebox{0.7}{
\begin{tabular}{llrrrrrrrr}
\toprule
\multirow{2}{*}{\textbf{Dataset}} &
\multirow{2}{*}{\textbf{Event Log}} & \multirow{2}{*}{\textbf{trace\#}} &  \multirow{2}{*}{\textbf{variant\#}} &  \textbf{event} &  \textbf{trace} &   \textbf{avg.\ trace} & \textbf{prefix}& \textbf{labeling} & \textbf{positive}  \\
& &  &   &  \textbf{class\#} &  \textbf{att.\#} &  \textbf{length} & \textbf{lengths} & \textbf{formula} & \textbf{class \%} \\
\toprule
BPIC2012\_{1}  &   \multirow{3}{*}{BPIC2012}  &      \multirow{3}{*}{4685} &  \multirow{3}{*}{3790} &               \multirow{3}{*}{36} &  \multirow{3}{*}{1}  & \multirow{3}{*}{35}& \multirow{3}{*}{[20,25,30,35]} & $\phi_1$ &  47\%\\
BPIC2012\_{2}  &          &     &       &                         &  &  & & $\phi_2$ & 17\%\\
BPIC2012\_{3}  &     &       &    &                 &                    &      &  & $\phi_3$ & 35\% \\ \hline                
BPIC2015\_1     &  BPIC2015\_1          &           696 &             677 &               380 &                      15 &                      42 & \multirow{5}{*}{[15,20,25,30]} &$\phi_4$ & 23\% \\
BPIC2015\_2    & BPIC2015\_2               &           753 &             752 &               396 &                      15 &                       55 & &$\phi_4$ & 19\% \\
BPIC2015\_3    & BPIC2015\_3               &          1328 &            1285 &               380 &                      15 &                       42 &  &$\phi_4$ & 20\% \\
BPIC2015\_4  & BPIC2015\_4                  &           577 &             576 &               319 &                      15 &                       42 &  &$\phi_4$ & 16\% \\
BPIC2015\_5    & BPIC2015\_5               &          1051 &            1049 &               376 &                      15 &                       50 & &$\phi_4$ & 31\% \\ \hline
BPIC2017\_{1} & \multirow{3}{*}{BPIC2017} & \multirow{3}{*}{31413} & \multirow{3}{*}{2087} & \multirow{3}{*}{36} & \multirow{3}{*}{3} & \multirow{3}{*}{35} & \multirow{3}{*}{[20,25,30,35]} &$\phi_5$  & 41\% \\ 
BPIC2017\_{2} &  &  &  &  &  & &  &$\phi_6$ & 12\% \\
BPIC2017\_{3} & &  &  &  &  & &  &$\phi_7$ & 47\% \\ \hline
Sepsis\_{1}  & \multirow{3}{*}{Sepsis} &   \multirow{3}{*}{782} &  \multirow{3}{*}{709} &   \multirow{3}{*}{15} & \multirow{3}{*}{24} & \multirow{3}{*}{13} & \multirow{3}{*}{[7,10,13,16]} &$\phi_8$ & 14\% \\
Sepsis\_{2}   &      &   &  &  &     &                      & & $\phi_9$ & 14\% \\
Sepsis\_{3}   &            &            &              &                 &                   &                      &   &$\phi_{10}$ & 14\% \\
\hline
Legal Complaints & Legal Complaints &  1101 & 636 & 14 & 4 & 2 &  [4,6,8,11]  &$\phi_{11}$ & 5\% \\
\bottomrule
\end{tabular}}
\caption{Event log statistics and labellings}
\label{tab:datastat}
\end{table}

The chosen datasets exhibit different characteristics, which can be seen in Table~\ref{tab:datastat}.
The smallest log is BPIC15\_4, which contains 577 cases, while the largest one is BPIC2017 with 31\,413
cases. The most imbalanced class labels are found in the  
Legal complaints dataset, where only $5\%$ of the cases are labelled as positive. 
Conversely,
in BPIC2012 and BPIC2017, the classes are almost perfectly balanced. In terms of event classes, 
the Legal complaints dataset is the one with the smallest number, with only 14 distinct activity classes. The ones with the largest activity alphabet size (number of different event classes), are, instead, 
the BPIC2015 datasets, reaching 396 event classes in BPIC2015\_2. The datasets also
differ in terms of the number of static attributes, i.e., trace attributes, ranging from a single trace attribute for BPIC2012 to $24$ trace attributes in Sepsis. Furthermore, an important aspect to consider is the trace-to-variant ratio. This statistic characterises the level of variability observed in the event log. The higher the number of variants to traces is, the more heterogeneous 
an event log is considered to be.  This, in turn, will also affect the results reported in the evaluation across the different datasets.

\subsection{Evaluation metrics}
\label{sec:metrics}
 
To evaluate the performance of the selected counterfactual approaches, 
we make use of the evaluation protocol presented in~\cite{caisepaper},~composed of seven evaluation metrics.

The chosen metrics refer to a single query instance \textit{x} to be explained and consider the returned counterfactual set $C = f_k(x,b,X)$.
Three of the seven metrics considered for the evaluation, that is, \emph{distance} ($o_2$), \emph{sparsity} ($o_3$), and \emph{implausibility} ($o_4$) are the same ones used for three of the four objectives described in Section~\ref{sec:c_general} and of the five objectives of the PPM approaches described in Section~\ref{sec:c_PPM}. Moreover, a fourth metrics, the \emph{trace fitness}, is the same one specifically introduced for the fifth objective of the PPM fitness function formulation (Section~\ref{sec:approach}). In other words, these four metrics are used to evaluate three or four dimensions optimised through the GA counterfactual generation approach for the baseline and adapted algorithms, respectively. At the end of the counterfactual generation process, we check that the predicted outcome of the counterfactuals is alway the desired outcome, as such we do not leverage the metrics related to the objective $o_1$ for the evaluation, as it is always $1$ for all the generated counterfactuals. Besides these, we also measure for the evaluation (i) a fifth quality metrics (the \emph{diversity}), (ii) the
\emph{hit rate}, which focuses on the capability of the methods to return the requested number of counterfactuals ($k$), as well as (iii) the \emph{runtime}, which measures the time required to generate the counterfactuals. 
For completeness, we shortly summarize in the following the considered evaluation metrics: 
\begin{itemize}[label=$\bullet$]
\item The \emph{distance} metrics measures the average distance between the query instance \textit{x} and the counterfactuals in \textit{C}. 

For good counterfactuals, the \emph{distance} metrics is low. This metrics is the same used for measuring the objective $o_2$.

\item The \emph{sparsity} metrics measures how many features have been changed in each counterfactual with respect to the query instance, averaged over the total number of counterfactuals generated. A low number of changed features is preferred. This metrics is the same used for measuring objective $o_3$.

\item The \emph{implausibility} metrics determines how close the generated counterfactuals are to the reference population. 
For good counterfactuals, the \emph{implausibility} metrics is low. This metrics is the same used for measuring the objective $o_4$.

\item The \emph{trace fitness} metrics determines the ratio of \declare constraints underlying the considered process that are satisfied in the query instance $x$ that are also satisfied by each counterfactual in $C$, averaged over the set of all generated counterfactuals. 
A high average \emph{trace fitness} indicates a high compliance of the generated counterfactuals with the constraints. This metrics is the same used for measuring the objective $o_5$.

\item The \emph{diversity} metrics measures the average pairwise \emph{distance} between the counterfactuals in $C$. The \emph{diversity} in $C$ should be high.

\item The \emph{hit rate} metrics is defined as $h/k$, where $h=|C|$ is the number of generated counterfactuals, and $k$ the number of requested counterfactuals (see Section \ref{sec:expl}). The value of \emph{hit rate} should be high. 

\item The \emph{runtime} (measured in seconds) is the time to compute the requested number of counterfactuals.

\end{itemize}

\subsection{Experimental setting}
\label{sec:setting}

To compare the four counterfactual generation approaches, we carried out different experiments with different datasets, trace prefix lengths, as well as different number of requested counterfactuals. 

The experiments were performed over traces with variable prefix lengths to determine whether different prefix lengths affect the results of the counterfactual generation process. We also performed experiments on both the simple-index and simple-trace index encodings introduced in Section~\ref{ch:ppmback}.
This was done (similarly to other PPM evaluations) to determine how the technique performs based on the amount of information available up to a certain cut-off point. The prefix lengths used for each dataset are based on the results of~\citet{irene}, where the selected prefix lengths ensure that the predictive model can perform adequately and provide good predictive performance. They are reported in Table~\ref{tab:datastat}.
We also examined different settings for the counterfactual generation process, where we ask for 5, 10, 15 and 20 counterfactual examples to be returned~\cite{guidotti2022counterfactual}. This was done to determine whether the number of counterfactuals has an impact on the quality of the generated counterfactuals, execution time and satisfiability of the counterfactuals. 
In the following we report first some details on the experimental procedure and then the specific settings related to the genetic algorithms.

\textbf{Experimental procedure.} 
For each dataset, trace encoding, 
prefix length, We split the data into $ 70 \% - 10 \% - 20 \%$ partitioning it into training, validation, and testing, respectively, using a sequential order split. A Random Forest (RF) model was trained using the Python library \texttt{scikit-learn}. For the predictive model, hyperparameter optimisation was performed using the \texttt{Hyperopt} Python library to identify the best model configuration for each dataset, prefix length and encoding combination. 
The training set used to train the RF model was used as input for the counterfactual generation methods to obtain $\mathcal{X}$.
Since we did not have knowledge on the process models generating the different logs, the \declare process constraints used as \bk, as well as for  computing the \emph{trace fitness} metrics in the evaluation, 
were discovered from the traces in the event log. This was done by considering all the \declare constraints satisfied by the  
traces in the whole training event log. To discover these constraints, a support threshold of 90\% was chosen for all datasets, except for the Sepsis datasets where a level of support of 99\% was chosen. This choice was made as it provided the best trade-off between the number of constraints discovered from each dataset and the ratio of non-conformant traces.\footnote{We used the \texttt{Declare4Py} Python Package~\href{https://github.com/ivanDonadello/Declare4Py/}{d4py}~\cite{declare4py} for the discovery of the \declare constraints.} The non-conformant traces are then discarded from the datasets, 
to ensure that all traces are compliant to the \emph{temporal background knowledge}. Afterwards, fifteen instances were sampled from the testing set and used for the counterfactual generation, one test set instance $x$ a time, where $x$ is a prefix of the trace. We then evaluate the generated counterfactuals using the evaluation framework presented in Section~\ref{sec:metrics}. All experiments were performed on a M1 Macbook Pro machine with 16GB RAM~\footnote{The code used to run the experiments can be found at~\href{https://github.com/abuliga/nirdizati_light_counterfactuals/}{repository}}.

\textbf{GA setting.}
For all four settings, the \textbf{initial population} is created through a hybrid procedure, either by finding points from the training set close to our test set instance $x$ or by randomly sampling values from the domain of the attributes.

In the \textbf{single-objective setting},  
the algorithm presented in~\citet{geco} is instantiated to find the most suitable counterfactual explanations. 
Since the PLAF specifications do not allow us to constrain
control-flow categorical variables,
they are not used in this setting. We make use of the implementation available within DiCE~\cite{dice}. 

For \boso, we make use of the baseline crossover and mutation operators available within DiCE, and the single-objective baseline fitness function presented in Eq.~\ref{eq:sobjbase}.
For \aoso, instead, we use the adapted crossover and mutation operators presented in Section~\ref{sec:adaptedoperators}, and the single-objective adapted fitness function reported in  
Eq.~\ref{eq:sobjadapted}.
The values of $\alpha$, $\beta$, $\gamma$ from Eq.~\ref{eq:sobjbase} are set to $\alpha = 0.5, \beta = 0.5, \gamma = 0.5$, the default configuration in DiCE. For \aoso, in Eq.~\ref{eq:sobjadapted}, the values for $\alpha$, $\beta$, $\gamma$ are the same, while the value of $\delta$ is set to $\delta = 0.5$, to balance the different objectives.

Regarding the \emph{population selection}, for both the single-objective approaches, at each iteration, 
the top 50\% of the old population is propagated to the new generation.
For the algorithm, a predefined number of generations is set to $100$ along with a performance threshold from one generation to another. If the number of total iterations is reached or a considerable improvement is not observed over the last generation, the algorithm terminates. 

In the \textbf{multi-objective setting}, we make use of the framework presented in~\citet{moc}, but we replace the multi-objective optimisation algorithm with the Adaptive Geometry Estimation-based Many Objective Evolutionary Algorithm (AGE-MOEA)~\cite{agemoea} to find the non-dominated counterfactual candidates.\footnote{We used the implementation available in the Python package~\href{https://github.com/anyoptimization/pymoo}{pymoo}~\cite{pymoo}.} 

AGE-MOEA can provide a set of non-dominated solutions that are well-distributed and that cover the entire hyperbolic surface of the objective space~\cite{agemoea}. Thus, it becomes an ideal algorithm to be used for generating a set of diverse counterfactual explanations, especially in a multi-objective setting. 

For the \bomo, we make use of the baseline crossover and mutation operators available in the implementation package, and the multi-objective baseline fitness function presented in Eq.~\ref{eq:mobjbase}.  
For \aomo, instead, we use the adapted crossover and mutation operators presented in Section~\ref{sec:adaptedoperators}, and the multi-objective adapted fitness function reported in Eq.~\ref{eq:mobjadapted}.

As for the \emph{population selection}, for both the multi-objective algorithms, at each iteration, the members of the next generation are selected from the non-dominated fronts, one front (or level) at a time. Then, AGE-MOEA selects the remaining solutions
from the last non-dominated front according to the descending order of their survival scores, computed as the distance from the neighbours and proximity to the ideal point.
The algorithm stops if either the predefined number of generations is reached ($100$) or
the performance no longer improves for a given number of successive generations.

\section{Results and Discussion}\label{sec:results}

\begin{figure}[tb]
\centering
 \includegraphics[width=\linewidth,height=12cm]{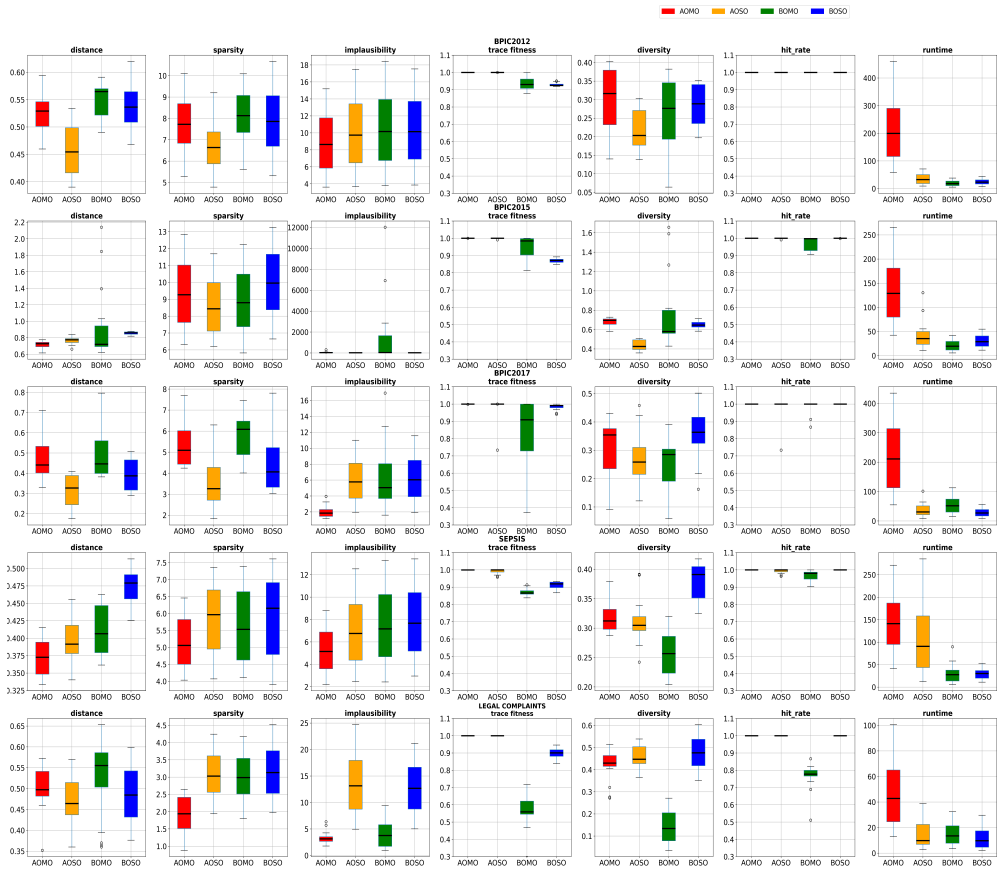}
 \caption{Results related to counterfactual quality, hit-rate, runtime and trace fitness for each event log.}
  \label{fig:overall_boxplots_logs}
\end{figure}

In this section, we aim at presenting the results of the experiments carried out and to answer the research questions introduced in Section~\ref{sec:eval}.
We report the overall results in \figurename~\ref{fig:overall_boxplots_logs}.
In the following, we first focus on \textbf{RQ1} (Section~\ref{sec:rq1}), we then address \textbf{RQ2} (Section~\ref{sec:rq2}), and we finally provide an overall discussion on the obtained results (Section~\ref{sec:discussion}).

\subsection{Results regarding RQ1}
\label{sec:rq1}

To answer \textbf{RQ1}, we compare the performance of the methods \aoso and \aomo, which make use of the single-objective or multi-objective adapted fitness functions together with the mating and crossover operators, with their baseline counterparts, that are, the \boso and \bomo methods. In Figure~\ref{fig:overall_boxplots_logs}, we report the overall results for the two trace encodings used, the different prefix lengths and the different number of requested counterfactuals.
By looking at \figurename~\ref{fig:overall_boxplots_logs}, we can make some general considerations concerning the impact of the inclusion of the adapted fitness functions and operators.

Overall, across the various datasets, the adapted counterfactual generation methods, namely \aoso and \aomo, consistently show either superior or equal performance compared to their baseline counterparts, \boso and \bomo. This conclusion holds across multiple evaluation metrics, especially in terms of conformance to the \bk (i.e., \tracefit) as expected, but also, in general, in terms of \distmetrics, \sparsity and \implmetrics. Indeed, by constraining the counterfactual generation process, we can generate counterfactuals closer to the original instance input (lower \distmetrics), less sparse, i.e., with a lower number of attributes changed, (lower \sparsity) and, in general, slightly more plausible (lower \implmetrics).

The only metrics for which the adapted versions do not outperform their baseline counterparts are \divemetrics and \runtime.
The difference in terms of \divemetrics is mainly attributable to the fact that we are constraining the control-flow execution during the generation process with the adapted operators, which we expected to have as an effect a reduction in \divemetrics. Actually, this is especially true for \aoso, which always has a lower average \divemetrics compared to the baseline method \boso, while \aomo always produces counterfactuals with a higher average \divemetrics compared to the baseline method \bomo. 

The difference in terms of \runtime is instead due to the conformance checks that are carried out when using the adapted fitness function, although the increase in \runtime varies from dataset to dataset. For \aomo the worst datasets in terms of runtime are BPIC2012 and BPIC2017, while for \aoso the worst dataset is Sepsis.

We also note that differences in dataset characteristics, such as outcome distribution and the number of event classes, impact the performance and \runtime of the methods. For instance, skewed distributions of the outcome labels and low event class numbers may contribute to longer convergence times or impact the quality of generated counterfactuals. In datasets that exhibit these behaviours, i.e., the Sepsis or Legal Complaints dataset, the differences in \runtime between the adapted methods and the baseline methods are less pronounced than in the other datasets, i.e., BPIC15, BPIC2012, BPIC2017.

In conclusion, the adapted methods \aoso and \aomo consistently outperform baseline methods \boso and \bomo in generating counterfactual instances across most of the evaluation metrics, except for a lower average \divemetrics for \aoso and a higher \runtime for both the adapted metrics (\textbf{RQ1}).

\subsection{Results regarding RQ2}
\label{sec:rq2}

To answer \textbf{RQ2}, we compare the performance of the methods leveraging a single objective fitness function (\aoso and \boso) with the multi-objective methods (\aomo and \bomo).

By looking at the plots in Fig.~\ref{fig:overall_boxplots_logs}, we can 
observe some notable differences.
In general, we can observe that \boso outperforms \bomo in several metrics. It is, in general, closer to the original query, less sparse (except for BPIC2015), more plausible (except for Legal Complaints), more diverse (except for comparable results for the BPIC2012 and BPIC2015 datasets), more compliant to the \bk (except for the BPIC2015 datasets), although slightly more costly in terms of \runtime. 

On the other hand, when looking at the adapted fitness function and operators, \aomo outperforms \aoso in terms of \distmetrics (except for BPIC2012 and BPIC2017), in terms of \implmetrics, \tracefit, \divemetrics, as well as \hr. In terms of \sparsity, instead, \aomo seems to perform better for more unbalanced datasets with a low number of event classes (Sepsis and Legal Complaints), while \aoso performs better on BPIC2012, BPIC2015 and BPIC2017. Finally, in terms of \runtime, \aoso significantly outperforms \aomo. 

By looking at these observations, we can conclude that the choice between \aoso and \aomo depends on specific priorities. \aoso may be preferred for applications where lower distance and sparsity are crucial, while \aomo might be more suitable for scenarios requiring higher diversity and where the runtime is not that important.

Overall, the different formulations of the fitness function in terms of single-objective or multi-objective optimization have an impact on the obtained results, however, the multi-objective optimization method benefits more from the inclusion of the adapted operators compared to the single-objective optimization, whereas for the baseline method, the single-objective \boso outperforms the multi-objective \bomo (\textbf{RQ2}).

\subsection{Discussion}
\label{sec:discussion}
Overall, we cannot identify a clear winner among the four investigated approaches. Indeed, although the adapted approaches outperform their counterparts in most of the metrics, the choice between \aoso and \aomo depends on specific priorities, such as the need for lower \runtime or higher \divemetrics, depending upon the requirements of the user.

To better understand other possible factors impacting the results, we also investigated and analysed the results by looking at different prefix lengths and different values of requested counterfactuals ($k$). 

\begin{figure}[tb]
\centering
 \includegraphics[width=\textwidth]{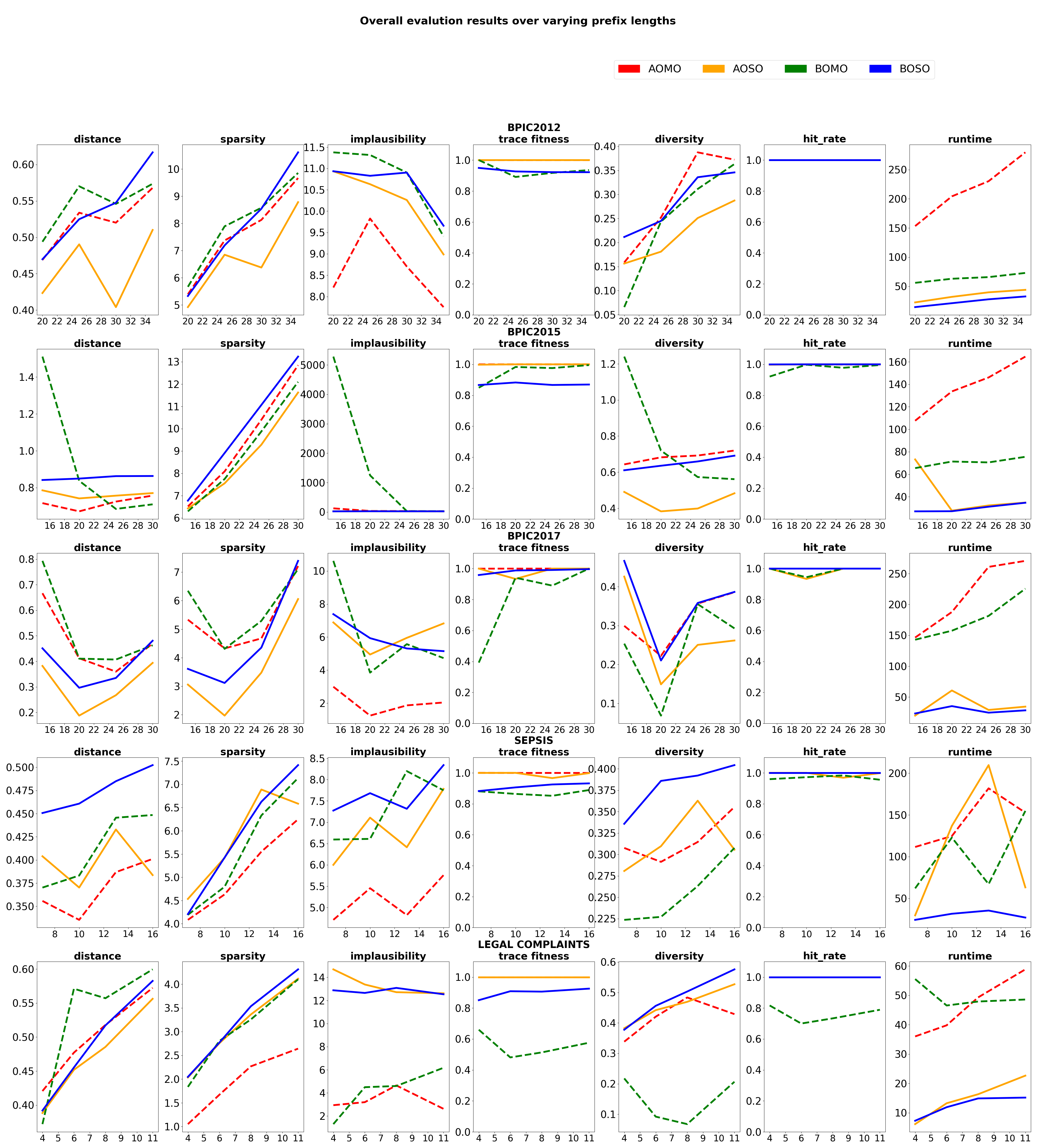}
 \caption{Results related to counterfactual quality, hit-rate, runtime and trace fitness for each event log, measured over variable prefix lengths.}
  \label{fig:avg_pl_logs}
\end{figure}

\figurename~\ref{fig:avg_pl_logs} reports the results of the evaluation varying the prefix lengths (according to the prefix length ranges defined in Table~\ref{tab:datastat}), where the results concerning different trace encodings and different datasets related to the same log are averaged.
The plot shows that the prefix length has an impact on the results. As a very general trend, we can observe that as the prefix length increases, the \distmetrics from the original instance, the \sparsity, the \divemetrics and the \runtime increase. However, the trend is not the same for all the datasets. Indeed, for some datasets (e.g., BPIC2012, BPIC2017) we can observe peaks for certain prefix lengths and anomalous behaviours and no clear trend as the prefix length increases. Additionally, we also note differences between the different datasets in terms of \implmetrics. While in some cases \implmetrics tends to decrease with longer prefixes, for other datasets (e.g., Sepsis), the opposite is true. This behaviour seems to depend on the dataset and the number of traces with longer prefix lengths. These points can be mainly explained by the heterogeneity of the traces of different lengths, and no clear trend can be identified by looking at the plot.

\begin{figure}[h]
\centering
 \includegraphics[width=\textwidth]{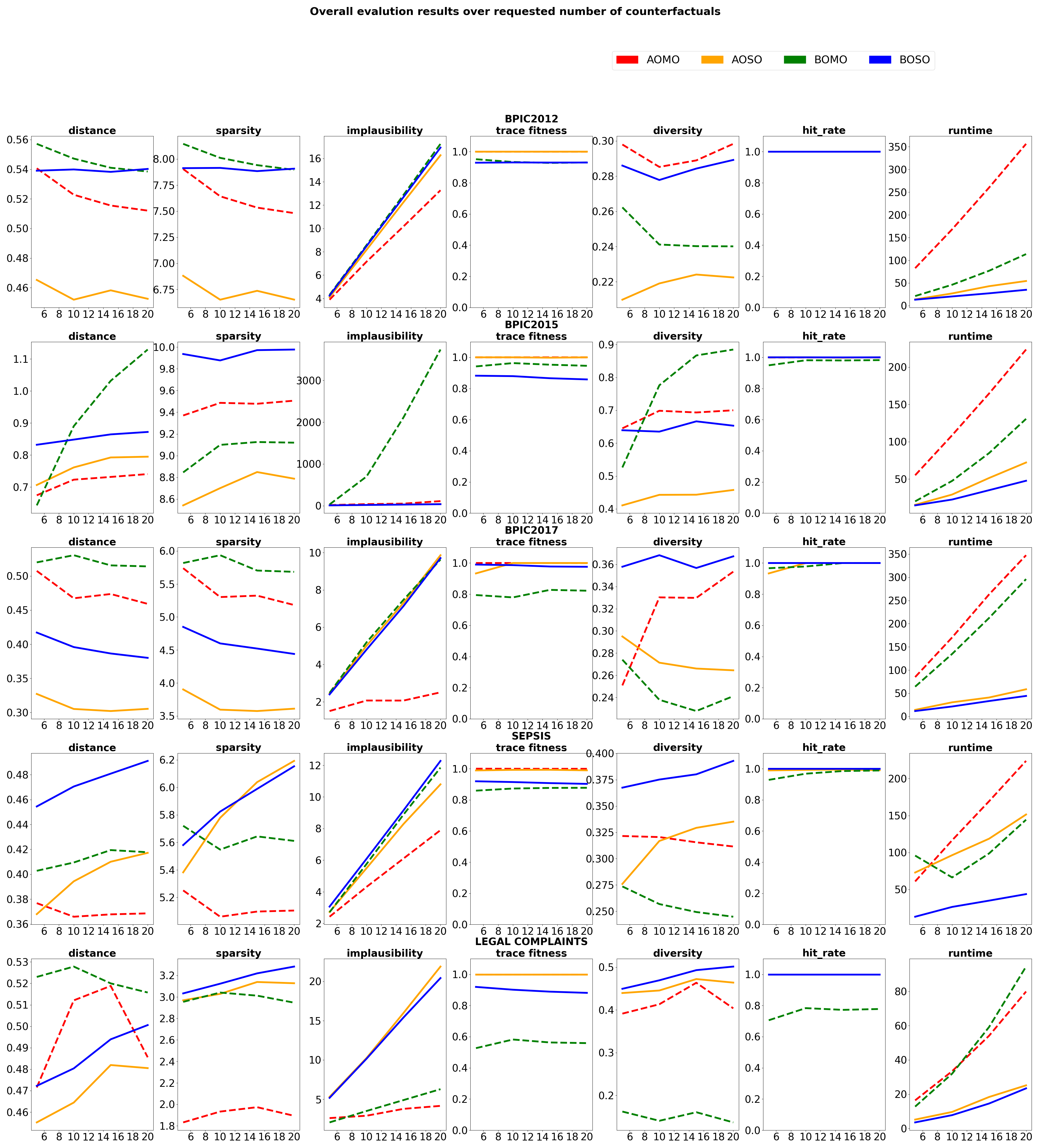}
 \caption{Results related to counterfactual quality, hit-rate, runtime and trace fitness for each event log, measured over an increasing number of requested counterfactuals.}\label{fig:avg_logs}
\end{figure}

Moreover, in line with the evaluation of~\citet{dice}, and~\citet{caisepaper}, we also show the results of the evaluation measured over an increasing number of requested counterfactuals $k$ in~\figurename~\ref{fig:avg_logs}. Also in this case, the results concerning different trace encodings, different datasets, and different prefix lengths related to the same log are averaged. The plot shows that in this case we can identify some clear trends. In particular, as expected, as the number of requested counterfactuals increases, the \implmetrics increases along with the \runtime.

Overall, in \figurename~\ref{fig:avg_pl_logs} and \figurename~\ref{fig:avg_logs}, we can see a correlation between the \implmetrics of the results and the \tracefit, especially for the \bomo method in the BPIC2017 and BPIC2015 datasets. As \implmetrics decreases, \tracefit increases, which confirms the fact that by using the adapted fitness function and genetic operators (i.e., the \aomo method), we can greatly improve both the \implmetrics and the \tracefit of the generated counterfactuals at the same time.
Moreover, we can also observe a strong correlation among \distmetrics and \sparsity metrics, which is especially clear for those datasets characterized by a large ratio of numerical trace attributes (e.g. BPIC2015, BPIC2012).

\section{Conclusions and Future work}
\label{sec:conclusions}

In this paper, we introduced a framework for generating counterfactual explanations with the inclusion of \bk at runtime for Genetic Algorithms (GAs). We first provided a systematic mapping from desiderata to concrete optimisation objectives for counterfactual generation using GAs. We then formalised the problem of generating counterfactual explanations through both a single-objective and multi-objective problem formulation for Predictive Process Monitoring. An adapted GA-based optimisation method was introduced that maintains the satisfaction of the \bk. The adapted method was applied for both the single-objective and multi-objective problem formulation and compared to state-of-the-art counterfactual GA approaches.

The results of the evaluation show that the adapted approaches that incorporate the \bk produce more conformant results, while also being on average less distant from the original instance and more plausible with the data manifold. However, as expected, this comes at the cost of an increase in complexity.

In the future, we would like to extend the constraints used in the approaches to also include the data-aware variants of such constraints, where other attributes are included together with the \textsc{Declare} constraints. Moreover, we would like to further refine our method to reduce the gap in terms of runtime required to return the generated counterfactual explanations. Finally, we would like to integrate \bk also for gradient-based optimisation techniques for the generation of counterfactual explanations.

\bibliography{sn-bibliography}

\end{document}